%% file: main.tex
\pdfoutput=1
\documentclass[11pt]{article}

\usepackage[preprint]{acl}
\usepackage{makecell}
\usepackage{times}
\usepackage{latexsym}

\usepackage[T1]{fontenc}
\usepackage[utf8]{inputenc}
\usepackage{times}
\usepackage{latexsym}
\usepackage{microtype}
\usepackage[noupquote]{inconsolata}

\usepackage{amsmath}
\usepackage{amsfonts}
\usepackage{amssymb}

\usepackage{graphicx}
\usepackage{booktabs}
\usepackage{array}
\usepackage{tabularx}
\usepackage{multirow}
\usepackage{pbox}
\usepackage{boxedminipage}
\usepackage{siunitx}
\usepackage{subcaption}
\usepackage{dblfloatfix}

\usepackage[table]{xcolor}

\usepackage{enumitem}
\usepackage{listings}
\usepackage{xstring}
\usepackage{needspace}
\usepackage{soul}
\usepackage{comment}
\usepackage{verbatim}

\usepackage{pifont}
\usepackage{fontawesome}

\providecommand{\todo}[2][]{}

\usepackage{hyperref}

\usepackage{xcolor}
\usepackage{listings}

\lstdefinestyle{smalllisting}{
  basicstyle=\small\ttfamily
}

\lstdefinestyle{aclprompt}{
  basicstyle=\ttfamily\scriptsize,
  columns=fullflexible,
  breaklines=true,
  breakatwhitespace=true,
  showstringspaces=false,
  keepspaces=true,
  frame=single,
  framerule=0.4pt,
  framesep=3pt,
  rulecolor=\color{black!25},
  xleftmargin=0pt,
  xrightmargin=0pt,
  framexleftmargin=0pt,
  framexrightmargin=0pt,
  linewidth=\columnwidth
}

\newcommand{\prompttitle}[1]{%
  \par\smallskip
  \noindent\textbf{#1}%
  \par\nobreak\smallskip
}

\definecolor{lightblue}{rgb}{.50,.90,0.51}
\definecolor{tri}{rgb}{.25,.88,.82}
\definecolor{lilac}{rgb}{0.85,0.64,0.85}
\definecolor{atomictangerine}{rgb}{1.0,0.6,0.4}

\newcommand{\benchmark}{\textsc{AHA-Memes}}

\title{%
  \textsc{AHA-Memes}: A Fine-Grained Multimodal Benchmark for
  Understanding Hate in Arabic Memes\\[0.4em]
  {\footnotesize\textcolor{red}{%
  Warning: This paper contains examples that may be disturbing to readers.}}%
}

\author{
Mohamed Bayan Kmainasi$^1$,
Ali Ezzat Shahroor$^1$,
Abul Hasnat$^2$,\\ 
\textbf{  
Md. Rafiul Biswas$^3$,
Wajdi Zaghouani$^4$,
Firoj Alam$^1$
}\\
$^1$Qatar Computing Research Institute, Qatar, 
$^2$APAVI.AI, France\\
$^3$Hamad Bin Khalifa University, Qatar, 
$^4$Northwestern University in Qatar, Qatar\\
\texttt{\{mkmainasi, alsh34060, mbiswas, fialam\}@hbku.edu.qa}\\
\texttt{mhasnat@gmail.com, wajdi.zaghouani@northwestern.edu}
}

\begin{document}
\maketitle
\begin{abstract}
Hateful memes are a growing form of multimodal online harm, where hostile intent is often conveyed through the joint interpretation of images, text, cultural references, and implicit targets. While hateful meme detection has advanced in high-resource languages, Arabic remains underexplored, with existing meme resources focusing mainly on propaganda or coarse harmful-content labels. We introduce \textsc{AHA-Memes},\footnote{AHA-Memes: \textbf{A}rabic \textbf{HA}teful \textbf{Memes}} which is, to our knowledge, the first large-scale Arabic hateful meme benchmark with fine-grained, multi-label annotations. The dataset includes 5K manually annotated memes using a taxonomy that captures hate types, i.e., attack strategies. We further provide $\sim$66K silver-labeled memes to support future studies. We benchmark text-only, image-only, and late-fusion multimodal models, as well as few-shot in-context learning (ICL) and open- and closed-weight Vision-Language Models (VLMs) under zero-shot and fine-tuning settings. Our results establish strong baselines and highlight key challenges in culturally grounded Arabic hateful meme detection. We release the dataset, annotation guidelines, and evaluation scripts to support future research.
\footnote{
Project resources:
\href{https://github.com/MohamedBayan/AHA-Memes}{GitHub}
\quad|\quad
\href{https://huggingface.co/datasets/QCRI/Arabic-Hateful-Memes}{Hugging Face}
}

\end{abstract}
 
\input{sections/introduction}
\input{sections/related_work}
\input{sections/dataset}
\input{sections/experiments}
\input{sections/results}
\input{sections/conclusion}

\section*{Limitations}
\benchmark{} is designed as a focused benchmark for Arabic hateful-meme understanding rather than a comprehensive representation of Arabic meme culture. We collected data four public platforms, Facebook, Instagram, Pinterest, and Twitter/X, which gives broad public coverage, however, our collection may not capture all dialects, regions, platform communities, or private and ephemeral content. Some annotation boundaries are inherently nuanced, especially between offensive humor, satire, political criticism, and protected-group hate. We address this detailed annotation guideline, and training. The auxiliary silver corpus broadens the resource and supports weakly supervised research, but its labels are model-generated and not human-verified. 

\section*{Ethics and Broader Impact}
\benchmark{} supports research on Arabic multimodal hate detection and safer content moderation. It is developed from publicly available memes and includes hateful, offensive, and sensitive material solely for research and evaluation. We do not collect or release personal information, which helps minimize privacy risks. As models can miss harmful content or over-flag legitimate content, they should be used with human oversight, clear documentation, and ongoing monitoring. Annotation was performed by trained, fairly compensated native-speaker annotators using guidelines that distinguish protected-group hate from generic offensiveness and legitimate criticism. To further reduce privacy and misuse risks, we release OCR text, labels, metadata, and image references.

\section*{Acknowledgments}
The work was supported by NPRP grant 14C-0916-210015 from the Qatar National Research Fund, part of the Qatar Research Development and Innovation Council (QRDI). The findings reported herein are solely the responsibility of the authors.

\bibliography{bibliography/main}

% \balance
\appendix

% \section{Example Appendix}
% \label{sec:appendix}

\input{sections/appendix}

\end{document}

%% file: sections/introduction.tex
\section{Introduction}
\label{sec:introduction}

\begin{figure}[t]
    \centering
    \includegraphics[width=0.46\textwidth]{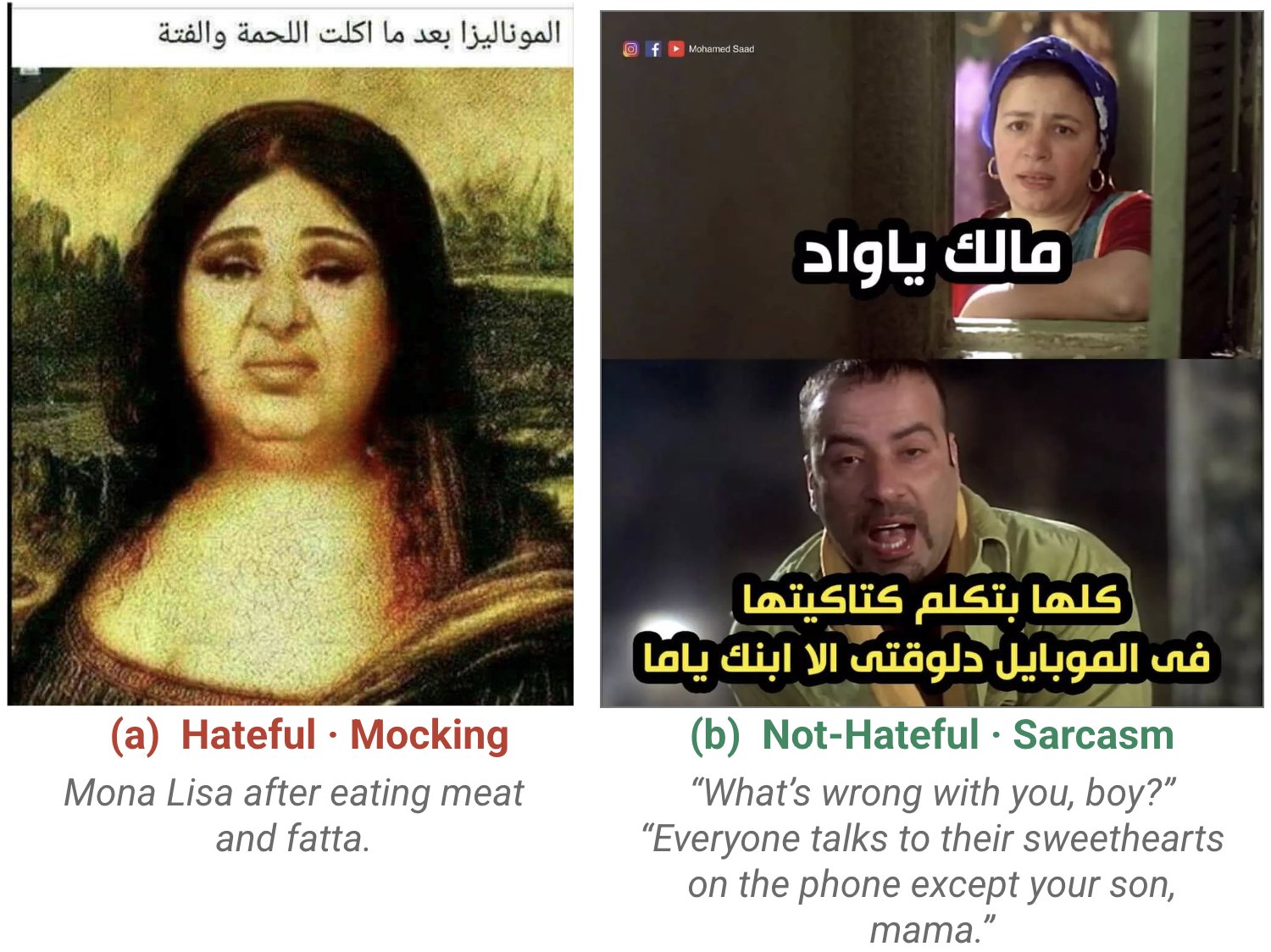}
    \vspace{-0.3cm}
\caption{Examples from the \benchmark{} dataset across hateful and non-hateful categories.}
\label{fig:examples}
\end{figure}

A smiling family photograph may appear harmless on its own, yet the same image can communicate hate when combined with a targeted caption. Likewise, a national flag paired with an apparently benign proverb may become sectarian or xenophobic depending on the communities, histories, and stereotypes that viewers bring to the meme. Memes therefore pose a distinctly multimodal and culturally grounded hate detection problem~\cite{shahroor2026memelens}. Their meaning is often produced through the joint interpretation of image, text, and shared background knowledge, rather than by either modality in isolation \cite{ijcai2022p781}. 
This dependence on cultural context also complicates human judgment, reduces label consistency, and makes automated detection more challenging. In a cross-country study, \citet{Bui20259714} found that annotators from five countries agreed on meme labels only 74\% of the time, with agreement falling to 67\% between U.S. and Indian annotators. Such disagreement among human annotators suggests that models trained mainly on culturally homogeneous and English-centric resources are likely to struggle when hate is expressed through culturally specific visual and linguistic cues. Figure~\ref{fig:examples} illustrates this with two examples from \benchmark{}: a\textit{ doctored portrait that mocks a woman's appearance (\emph{Hateful}), and a television still repurposed as everyday \emph{sarcasm} (\emph{Not-Hateful}), where the hateful or benign reading emerges only from the image--text pairing}.

Despite growing interest in multimodal hate detection, existing meme benchmarks remain focused in English and often rely on binary or coarse labels~\cite{Kiela2020}. This limits their ability to capture \textit{who is targeted}, \textit{what form of hate is expressed}, and \textit{how hateful meaning is constructed} through the interaction of image, text, and cultural context. Recent multilingual efforts have begun to broaden coverage, but available resources remain limited in scale and granularity. For example, Multi3Hate provides parallel multilingual annotations but only 300 samples per language \cite{Bui20259714}, while GuardHarMem introduces fine-grained meme hate labels only for English \cite{El-amrany2025}. Arabic is especially underserved in this space. Although Arabic text-based hate-speech detection has advanced through shared tasks such as OSACT \cite{ Mubarak2023OSACT5}, and Arabic meme research has studied propaganda in ArMeme \cite{alam2024propaganda, kmainasi-etal-2025-memeintel}, there is still no dedicated benchmark for fine-grained hate detection in Arabic memes. 
% Mubarak2020OSACT,
% alam2022overview, 
% This gap is important because Arabic memes combine right-to-left embedded text, dialectal variation, code-switching, and culturally specific political, sectarian, gendered, and national symbols. These factors make hate detection difficult for both annotation and modeling.

We address this gap with \benchmark{}, a manually annotated fine-grained multimodal dataset for Arabic hateful memes. The dataset includes hateful and non-hateful memes and provides multi-label annotations for hate types. Figure~\ref{fig:pipeline} gives an overview of the end-to-end construction pipeline.
% and targeted protected categories, enabling analysis beyond binary detection. 
We conduct detailed experiments on broad set of systems, including text encoders, image encoders, multimodal fusion models, open-weight VLMs, and closed-weight VLMs and LLMs. Our experiments show that fine-tuned open VLMs perform best for binary detection, that embedded text carries much of the signal while visual information remains complementary, and that fine-grained hate categorisation remains challenging. We also evaluate retrieval-augmented few-shot in-context learning as a training-free alternative.
Our contributions are as follows.
\begin{itemize}[noitemsep,topsep=0pt,leftmargin=*,labelsep=.5em]
\item We introduce \benchmark{}, the first large-scale Arabic hateful-meme dataset, consisting of \emph{5K manually annotated} memes with binary hate labels and fine-grained multi-label hate types. In addition, we provide silver labels\footnote{By silver labels, we refer to labels generated by an LLM. For this task we used Gemini-3.1-Pro.} for targeted protected categories and features such as dialect, emotion, and sentiment.

\item We develop an Arabic-contextual bilingual annotation scheme for binary hate labels, fine-grained hate types, and targeted protected categories, which can support future dataset development for Arabic hateful-meme analysis.

\item We provide a comprehensive benchmark across five model families, comparing text-only, image-only, multimodal fusion, open-weight VLM, and closed-weight VLM/LLM systems under zero-shot, few-shot, and fine-tuning settings.

\item We release an additional $\sim$66K Arabic meme corpus with silver labels for binary hate, fine-grained hate type, target category, dialect, emotion, and sentiment, enabling future work on large-scale Arabic multimodal harmful-content modelling.

\end{itemize}

%% file: sections/related_work.tex
\section{Related Work}
\label{sec:related_work}

\paragraph{Hateful Meme Detection and Resources.}
The Facebook Hateful Memes Challenge \cite{Kiela2020} established hateful-meme detection as a multimodal task that requires joint reasoning over image and text. Its use of benign confounders showed that neither modality alone is sufficient, motivating later work on cross-modal fusion, visual-to-text prompting, and LLM-based reasoning for meme interpretation \cite{kumar-nandakumar-2022-hate,Cao2022321,Lin20242359}. More recent studies further show that external knowledge, cross-modal alignment, and cultural grounding are important for reliable meme understanding \cite{ijcai2022p781,Ren2026,kmainasi-etal-2025-memeintel}. However, most research in this area remains English-centric, and recent evidence suggests that VLMs may inherit culturally narrow assumptions when applied to non-English memes \cite{wang2026native}.

Recent efforts have begun to address under-resourced languages, including Bengali, Tamil, Malayalam~\cite{shahroor2026memelens}.
% , Indonesian, Chinese, Korean, and Amharic. 
However, these resources remain fragmented in scale, language coverage, and annotation depth \cite{Das202315498,Ponnusamy20247480,Lu2024}. 
% The only multilingual parallel hateful-meme resource, Multi3Hate \cite{Bui20259714}, contains only 300 samples per ?language, covering 5 languages. 
In addition, most resources use binary or coarse labels, which obscure \emph{\textbf{who}} is targeted, \emph{\textbf{what}} type of hate is expressed, and \emph{\textbf{how}} hate is rhetorically framed. 

\begin{figure*}[t]
    \centering
    \includegraphics[width=0.90\textwidth]{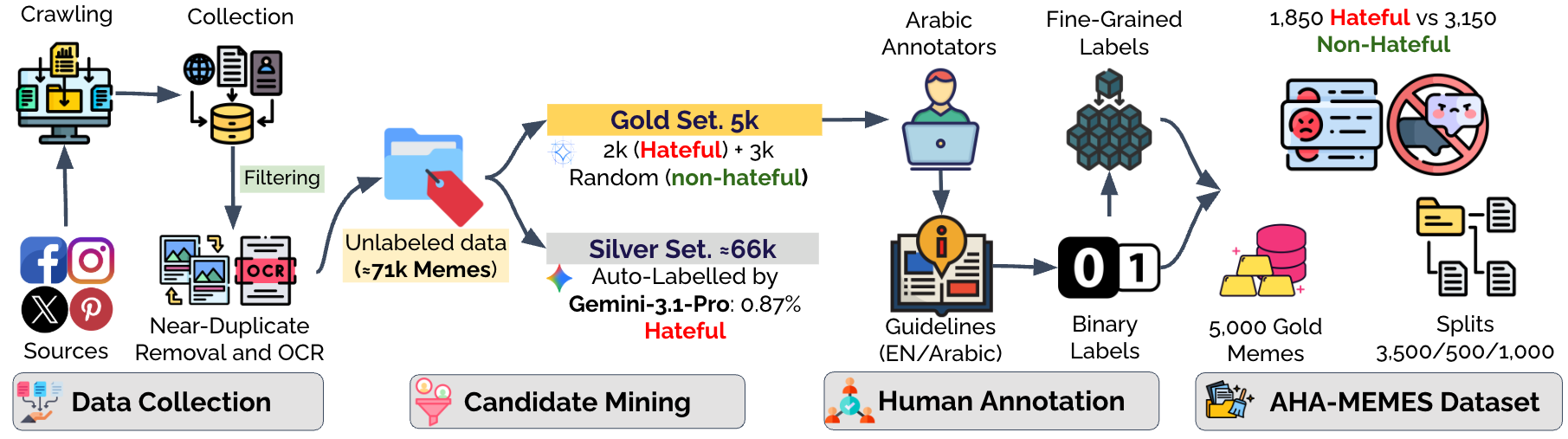}
    \vspace{-0,2cm}
    \caption{Overview of the \benchmark{} construction pipeline.
    % : \textbf{Data Collection} from public social platforms with near-duplicate removal and OCR ($\sim$71K memes); \textbf{Candidate Mining}, where a Gemma-3-12B pre-filter yields a 5K \emph{gold} set for annotation and a $\sim$66K Gemini-3.1-Pro \emph{silver} set; \textbf{Human Annotation} into binary and fine-grained labels; and the resulting \textbf{\benchmark{}} dataset (1{,}850 Hateful / 3{,}150 Not-Hateful; splits 3{,}500/500/1{,}000).
    }
    \vspace{-0,2cm}
    \label{fig:pipeline}
    \vspace{-0.3cm}
\end{figure*}

% The dataset landscape reflects this imbalance. 
% English resources cover a wide range of meme-related tasks, including hateful, harmful, offensive, misogynistic, propaganda, and sentiment-oriented content \cite{Kiela2020,Gomez2020MMHS150K,Fersini2022MAMI}. 
% Compared to that Non-English resources have begun to appear for languages such as Bengali, Tamil, Malayalam, Indonesian, Chinese, Korean, and Amharic, however, they remain fragmented in scale, language coverage, and annotation depth \cite{Das202315498,Ponnusamy20247480,Lu2024}. The only multilingual parallel hateful-meme resource, Multi3Hate \cite{Bui20259714}, contains only 300 samples per language. In addition, most resources use binary or coarse labels, which obscure \emph{who} is targeted, \emph{what} type of hate is expressed, and \emph{how} hate is rhetorically framed. 
% Fine-grained efforts such as GuardHarMem \cite{El-amrany2025} and ToxiCN-MM \cite{Lu2024} are important steps in this direction, however, no dedicated benchmark provides fine-grained hateful-meme annotation for Arabic.

\paragraph{Arabic Resources.}

Arabic multimodal disinformation or harmful-content research has been driven mainly by text-based datasets and shared tasks. OSACT advanced offensive-language and hate-speech detection in Arabic social media \cite{Mubarak2023OSACT5}, while related shared tasks and datasets have addressed propaganda detection and explanation in Arabic and multilingual settings \cite{hasanain-etal-2025-propxplain}. Recent multilingual and multimodal research on Arabic has addressed fact-checking \cite{alam2023overview} and the understanding of news and social-media content \cite{kmainasi-etal-2025-llamalens}.
Arabic multimodal social-media analysis has focused primarily on propaganda rather than hate. The ArMeme dataset~\cite{alam-etal-2024-armeme} and ArAIEval shared tasks \cite{hasanain-etal-2024-araieval} introduced Arabic meme propaganda-technique detection, followed by work on coarse hate labels, explanation-enhanced modelling, and explainable Arabic-English propaganda detection \cite{alam2024propaganda,kmainasi-etal-2025-memeintel,alam2022overview}. These resources are valuable for studying persuasion and propaganda, but they do not provide systematic fine-grained annotations for hateful memes, including hate, and fine-grained hate type, as presented in Table \ref{tab:arabic-meme-datasets}. We address this gap by developing \benchmark{}, a dedicated fine-grained benchmark for Arabic hateful-meme detection.

\begin{table*}[t]
\centering
\setlength{\tabcolsep}{2pt} 
\scalebox{0.68}{
\begin{tabular}{lcccll}
\toprule
\textbf{Dataset} & \textbf{Lang.} & \textbf{Size} & \textbf{Multi-label} & \textbf{Task granularity} & \textbf{Source} \\
\midrule
ArMeme \citep{alam-etal-2024-armeme} & AR & 5,725 & No & Propaganda status & Social media \\
ArAIEval \citep{hasanain-etal-2024-araieval}$^{\dagger}$ & AR & 3,062 & No & Binary propaganda & ArMeme \\
Propaganda-to-Hate \citep{alam2024propaganda}$^{\dagger}$ & AR & $\sim$3.1K & Yes & Hate + hate types & ArAIEval \\
MAHED Task~3 \citep{zaghouani-etal-2025-mahed}$^{\dagger}$ & AR & 3,562 & No & Binary hate & Social media \\
MemeXplain \citep{kmainasi-etal-2025-memeintel}$^{\dagger}$ & AR/EN & 5,725 AR + 10K EN & No & Label + explanations & ArMeme + Hateful Memes \\
AraHarMeme \citep{elamrany-etal-2026-araharmeme} & AR & 5,313 & No & Three-way harmfulness & Social media \\
ArPoMeme \citep{zaghouani-etal-2026-arpomeme} & AR & $\sim$7.3K & Yes & Ideology and polarization & Political memes \\
\midrule
\textbf{AHA-MEMES (ours)} & \textbf{AR} & \textbf{5K + 66K$^{\ddagger}$} & \textbf{Yes} & \textbf{Hate + 8 hate / 3 non-hate types} & \textbf{four platforms} \\
\bottomrule
\end{tabular}
}
\vspace{-0.2cm}
\caption{Comparison of Arabic multimodal meme datasets. $^{\dagger}$The resource contains data derived from or overlapping with an earlier Arabic meme dataset. $^{\ddagger}$AHA-Memes contains 5K human-annotated gold memes and approximately 66K auxiliary silver-labeled memes.}
\label{tab:arabic-meme-datasets}
\vspace{-0.3cm}
\end{table*}

% alam2022overview

% These efforts provide important foundations, however, they do not address the multimodal and culturally grounded nature of hateful memes.
%
% Mubarak2020OSACT,
% These resources are valuable for studying persuasion and propaganda, however, they do not provide systematic fine-grained annotations of hate type and target. 
% This distinction is important because hateful memes often require identifying both the protected group being attacked and the rhetorical form of the attack.

% Arabic memes also introduce linguistic and cultural challenges that are not well captured by English-centric resources. Right-to-left embedded text complicates optical character recognition, dialectal variation across Modern Standard Arabic (MSA) and regional varieties affects both extraction and interpretation, code-switching is common, and political, sectarian, gendered, tribal, national, and refugee-related symbols often require region-aware background knowledge. Since meme labels and safety judgments do not transfer cleanly across communities \cite{Bui20259714,wang2026native}, therefore, we design \benchmark{} to capture Arabic-specific linguistic phenomena.
% Arabic-specific cultural and linguistic phenomena, including sectarian hostility, anti-refugee rhetoric, tribal-national rivalry, and gender-based discrimination.

\paragraph{In-Context Learning for Multimodality.} % Classification}

In-context learning offers a training-free alternative to fine-tuning, however, its effectiveness depends strongly on which demonstrations are included in the prompt \cite{brown2020gpt3,abdelali-etal-2024-larabench}. Similarity-based selection is a common strategy for selecting useful examples in text classification \cite{liu2022makes}. 
% In multimodal learning, few-shot VLMs such as Flamingo \cite{alayrac2022flamingo} popularized ICL, while retrieval-based selectors such as RICES \cite{yang2022rices} use visual similarity to choose demonstrations.
%
In multimodal learning, most ICL work has focused on captioning and visual question answering, often using retrieval from a single modality. It remains underexplored for multimodal \emph{classification}, especially in culturally grounded and low-resource settings. We address this gap by selecting demonstrations through reciprocal-rank fusion of text and image similarities~\cite{cormack2009rrf}.

%% file: sections/dataset.tex
\section{Dataset}
\label{sec:dataset}

% \benchmark{} comprises 5{,}000 Arabic memes, each an image paired with its
% embedded text, collected from public social media and annotated with a
% hierarchical, conditional hate-speech scheme. We treat a \emph{meme} as an
% image--text pair whose meaning arises from the interaction of the two modalities.
% We describe collection (\S\ref{ssec:collection}), filtering and OCR
% (\S\ref{ssec:filtering}), the annotation scheme and workflow
% (\S\ref{ssec:annotation}), agreement (\S\ref{ssec:agreement}), and the resulting
% splits (\S\ref{ssec:statistics}); the full guidelines (English and Arabic) are in
% Appendix~\ref{sec:annotation_guidelines}.

\subsection{Data Curation}
\label{ssec:collection}
We sourced memes from Facebook, Instagram, Pinterest, and Twitter~(X). For Facebook, Instagram, and Pinterest, we manually selected public groups and pages focused on public figures, politics, and social commentary, contexts in which hateful memes are likely to circulate. We then collected images from these public sources using a semi-manual procedure following prior work \cite{alam-etal-2024-armeme}. For Twitter~(X), we collected memes through keyword-based search. In Figure \ref{fig:pipeline}, we provide the data curation pipeline. 

%
% \textcolor{red}{
% Since these platforms do not provide suitable public APIs, therefore, we used a semi-automatic pipeline that loads content in a browser and extracts the displayed images, following prior work \cite{alam-etal-2024-armeme}.
% \textit{HASNAT: I suggest to ignore this part on how we collected the data from social media as that type of collection can be questioned rather than considered as a standard procedure. Just refer to old work or totally ignore.}
% } 
%

% content was gathered via keyword-based
% retrieval of media posts using terms relevant to Arabic hate speech.

\subsection{Filtering and OCR Text}
\label{ssec:filtering}

\textbf{Duplicate Removal.} Social-media collections are highly redundant as users often repost identical or slightly modified content. We therefore removed exact and near-duplicate images after initial data curation. For near-duplicate filtering, we extracted image embeddings using a model fine-tuned on social-media images \cite{alam2020deep} following the approach discussed in \cite{alam-etal-2024-armeme}. We then computed pairwise Euclidean distances between image embeddings across the full dataset. Image pairs with a distance of $3.6$ or lower were treated as near-duplicates, and only one image from each duplicate cluster was retained.

% by encoding each meme with a ResNet18 embedding \cite{he2016deep} fine-tuned on social-media imagery-based model \cite{alam2020deep} and discarding any meme within a similarity threshold of $3.6$. of a
% retained neighbour under Euclidean distance. 

\noindent
\textbf{OCR Text.} We extracted embedded text using EasyOCR,\footnote{\url{https://github.com/JaidedAI/EasyOCR}} which we selected for its broad script coverage and scene-text detection capabilities \cite{liao2022real}. Because our study focuses on multimodal image-text analysis, we discarded memes with no detectable embedded text. As a result, every retained meme contains both visual and textual modalities.

% \noindent
% \textbf
\subsection{Image Selection for Annotation.}
Harmful, hateful, and misleading content represents a relatively small proportion of online content, however, its societal impact can be substantial, as reported in prior studies \cite{ijcai2022p781}. To increase the proportion of relevant examples while preserving coverage of non-hateful memes, we used a weak pre-selection strategy before manual annotation. Specifically, we applied Gemma-3-12B to assign preliminary binary hateful labels to 71K memes. We then selected 5K memes for manual annotation from this pre-selected pool. This sampling strategy produced a balanced and representative annotation set, consisting of 1{,}850 hateful memes and 3{,}150 non-hateful memes.
% alam-etal-2022-survey
% For the annotation tasks,  
% has been reported in myHateful memes can be rare online, so annotating a random sample might yield too few
% positives. We therefore used an open model, Gemma-3-12B, as a recall-oriented pre-filter, and built the $5{,}000$-meme set from $\sim$$2{,}000$ memes flagged as hateful, along with $\sim$$3{,}000$ randomly selected non-hateful memes. These predictions only prioritised memes for annotation: they were hidden from annotators and discarded, so all gold labels are human (\S\ref{ssec:annotation}). The random sample keeps a realistic negative class and recovers hate the filter missed; after annotation, $1{,}850$ memes are \emph{Hateful} and
% $3{,}150$ \emph{Not-Hateful}.

\subsection{Annotation}
\label{ssec:annotation}

\paragraph{Manual Annotation Tasks.}
We manually annotate each meme for binary hatefulness and, fine-grained hierarchical labels. The annotation scheme is organized as follows.

\begin{itemize}[noitemsep,topsep=0pt,leftmargin=*,labelsep=.5em]
\item \texttt{is\_hateful} is a binary judgment. Following established definitions \cite{Kiela2020}, a meme is labeled as \emph{Hateful} if it directly or indirectly attacks people on the basis of a protected characteristic, including ethnicity, race, nationality, immigration status, religion, caste, sex, gender identity, sexual orientation, disability, or disease. We make two boundary cases explicit. \textit{First}, attacks against groups that themselves promote or perpetrate hate, such as terrorist organizations, are not treated as hate speech. \textit{Second}, content that is offensive, rude, or insulting but does not target a protected category is labeled \emph{Not-Hateful}. This separates hatefulness from generic offensiveness.

\item For memes labeled \emph{Hateful}, \texttt{hateful\_type} captures the attack strategy as a multi-label annotation over \emph{Mocking}, \emph{Incitement}, \emph{Dehumanization}, \emph{Slurs}, \emph{Contempt}, \emph{Inferiority}, and \emph{Exclusion}. A meme may receive multiple hate-type labels (e.g., \emph{Slurs} and \emph{Dehumanization}).
% Annotators also identify the targeted protected category or categories.

\item For memes labeled \emph{Not-Hateful}, \texttt{non\_hateful\_type} assigns one or more labels from \emph{Humor}, \emph{Sarcasm}, and \emph{Other}. 
\end{itemize}

This hierarchical structure supports binary hate detection at the top level, enables fine-grained categorization for hateful content, and ensures that fine-grained labels are consistent with the binary decision by construction. We provide category definitions and examples in Appendix~\ref{sec:annotation_guidelines}.

\paragraph{Annotation Team.}
Annotation was conducted by a third-party company. Three trained native Arabic-speaking annotators annotated the dataset using bilingual guidelines (See Appendix \ref{sec:annotation_guidelines}) in English and Arabic, which we developed to capture dialectal and culturally specific cues \cite{hasanain2024can}. Before annotation, annotators were informed that the dataset may contain offensive, hateful, or otherwise harmful content. In addition, they were informed about the use of the data. We also conducted multiple rounds of training and guideline refinement to improve annotation consistency and quality. In total, \benchmark{} contains 5K manually annotated memes, with an annotation cost of $\sim\$4K$. 

% alam2021fighting
% \paragraph{Annotation team.}
% Annotation has been completed by a third party company. Three trained native-Arabic-speaking annotators annotated the dataset. We developed annotation guideline in bilingual annotation guidelines in English and in Arabic to capture dialectal
% and cultural cues \cite{alam2021fighting, hasanain2024can}. We also informed them about the type content they may see during the annotation phase. Multiple rounds of training were provided to ensure the higher quality of the annotation. In total, \benchmark{} contains 5{,}000 memes, which resulted in a cost of $\$$4K.

% and   working from guidelines authored in English and translated into Arabic to capture dialectal
% and cultural cues \cite{alam2021fighting, hasanain2024can}. 
% Annotation proceeded in two calibration rounds (each a triple annotation of the same 250 memes, 750 records per round) separated by a calibration session, followed by a single large-scale phase in which the remaining 4{,}500 memes were single-annotated and distributed evenly across annotators. 
% In total, \benchmark{} contains 5{,}000
% unique memes and 6{,}000 annotation records (500 triple-annotated $+$ 4{,}500
% single-annotated), with each annotator contributing 2{,}000 records.

\paragraph{Annotator Agreement}
\label{ssec:agreement}

We measured agreement with Cohen's $\kappa$ agreement on the annotated dataset. For each multi-label task, every subtype was treated as a separate binary label. Cohen's $\kappa$ was computed independently for each subtype and each of the three annotator pairs, then macro-averaged across subtypes and annotator pairs.
Annotation agreement are 0.91, 0.75, and 0.67 for \texttt{is\_hateful}, \texttt{hateful\_type}, and \texttt{non\_hateful\_type}, respectively, indicating substantial to near-perfect agreement for all annotation tasks \cite{landis1977measurement}.

\subsection{Statistics and Data Splits}
\label{ssec:statistics}

\paragraph{Distribution.}
\benchmark{} is moderately imbalanced, with 1{,}850 memes (37.0\%) labeled as \emph{Hateful}. The fine-grained label distribution is long-tailed (Table~\ref{tab:datasplit}). On the non-hateful side, \emph{Sarcasm} (1{,}393) and \emph{Humor} (1{,}331) are the most frequent labels. Among hateful memes, \emph{Mocking} (1{,}007) is the most common attack strategy, whereas \emph{Exclusion} (17) is the rarest. Since fine-grained labels are multi-label, the total label count exceeds the number of memes. The proportion of multi-label memes also increases from 9.9\% in the train split to 21.4\% in the test split, reflecting the richer co-occurring strategies captured in the triple-annotated gold subset. Per-split label cardinality is reported in Appendix~\ref{ssec:data_analysis}.

\paragraph{Splits.}
We partition \benchmark{} into train, development, and test splits containing 3{,}500 (70\%), 500 (10\%), and 1{,}000 (20\%) memes, respectively. The splits are stratified by the binary hate label, and we ensure that no meme appears in more than one split. The label distribution across the train, development, and test sets is reported in Table~\ref{tab:datasplit}.
% for both binary and fine-grained categories.

 % each retain a 37.8\% hateful rate, while the gold-enriched test set contains 33.7\% hateful memes (

% To improve evaluation reliability, we place all 500 triple-annotated gold memes in the test set and add 500 single-annotated memes, so that half of the test set has majority-verified labels. The train and development sets are drawn entirely from the single-annotated pool. We stratify splits by the binary hate label, remove meme overlap across splits, and use a fixed random seed of 42. The train and development sets each retain a 37.8\% hateful rate, while the gold-enriched test set contains 33.7\% hateful memes (Table~\ref{tab:datasplit}).

\begin{table}[]
\centering
\setlength{\tabcolsep}{2pt} 
\scalebox{0.85}{
\begin{tabular}{@{}llrrrr@{}}
\toprule
\multicolumn{1}{c}{\textbf{Hate}} & \multicolumn{1}{c}{\textbf{Fine-grained}} & \multicolumn{1}{c}{\textbf{Train}} & \multicolumn{1}{c}{\textbf{Dev}} & \multicolumn{1}{c}{\textbf{Test}} & \multicolumn{1}{c}{\textbf{Total}} \\ \midrule
Hateful & Mocking & 706 & 90 & 211 & 1,007 \\
Hateful & Incitement & 320 & 51 & 85 & 456 \\
Hateful & Dehumanization & 247 & 42 & 58 & 347 \\
Hateful & Slurs & 252 & 42 & 47 & 341 \\
Hateful & Contempt & 107 & 18 & 50 & 175 \\
Hateful & Inferiority & 57 & 14 & 32 & 103 \\
Hateful & Exclusion & 10 & 4 & 3 & 17 \\
Hateful & Other (H) & 18 & 2 & 7 & 27 \\ \midrule
Not Hateful & Other (NH) & 380 & 50 & 96 & 526 \\ 
Not Hateful & Sarcasm & 934 & 126 & 333 & 1,393 \\
Not Hateful & Humor & 863 & 136 & 332 & 1,331 \\ \bottomrule
\end{tabular}
}
\vspace{-0.2cm}
\caption{Fine-grained sub-type distribution across splits. Counts sum to more than the number of memes because the label is multi-label. \emph{Humor}, \emph{Sarcasm}, and \emph{Other} are non-hateful subtypes; the rest are hateful attack types. H: Hateful; NH: Not-Hateful.}
\label{tab:datasplit}
\vspace{-0.3cm}
\end{table}

\subsection{Auxiliary Silver Dataset}

In addition to the 5K human-verified memes that constitute the core \benchmark{}, we release an auxiliary dataset of $\sim$66K Arabic memes with LLM-generated annotations from Gemini~3.1~Pro.\footnote{\url{https://deepmind.google/models/model-cards/gemini-3-1-pro/}} We refer to this resource as a \emph{silver} dataset ~\cite{wang-etal-2024-use,du-etal-2023-effective} as its labels are not human-verified. The dataset is intended to support weakly supervised learning, semi-supervised learning, active learning, retrieval-based analysis, and future annotation expansion.

The motivation for releasing this auxiliary set is twofold. \textit{First}, even without labels, a large in-domain collection of Arabic memes is valuable for the community. It captures meme templates, OCR noise, dialectal and code-mixed text, culturally specific references, visual stereotypes, and rapidly evolving political and social topics. Such data can support meme-understanding tasks that require joint reasoning over image content, embedded text, social context, and target entities \cite{pramanick-etal-2021-momenta-multimodal}. \textit{Second}, the silver labels provide a scalable starting point for methods that explicitly model noisy supervision. Weak-supervision frameworks have shown that noisy heuristic labels can be useful when denoised or combined with small amounts of gold data \citep{ratner2016data,ratner2017snorkel}. Similarly, semi-supervised and self-training methods use pseudo-labels on large unlabeled collections to improve model learning, especially when combined with confidence filtering and consistency-based training \citep{sohn2020fixmatch}. Recent work on LLM-assisted annotation further suggests that LLMs can reduce annotation cost and help bootstrap labeled resources, while also emphasizing the importance of distinguishing model-generated annotations from human-verified labels %and evaluating their limitations 
\citep{tan-etal-2024-large,he-etal-2024-annollm}.

% We position the $\sim$66K set as an auxiliary community resource, while all primary results use the human-verified 5K set. Gemini~3.1~Pro generates labels, subtypes, and metadata for the unlabelled memes; for the gold set, it generates only metadata conditioned on the human labels and does not relabel them. The silver data indicate that hateful content is relatively rare, while the collection remains dialectally and topically diverse. Full distributions and prompts are provided in Appendix~\ref{ssec}.

We therefore position the $\sim$66K set as an auxiliary community resource rather than as a core contribution. All primary results in this paper are reported on the human-verified 5K set. Silver annotations are produced with Gemini~3.1~Pro using two prompts: for the unlabelled $\sim$66K memes the model predicts the label and subtype together with rich metadata (topic, dialect, sentiment, target categories, propaganda techniques, etc.), whereas for the 5K gold memes it generates the same metadata \emph{conditioned} on the human label, so it never relabels them. The silver labels indicate that hateful content is relatively rare while the collection remains dialectally and topically diverse. Full distributions and the prompts are provided in Appendix~\ref{ssec:silver_stats}.

% \todo[inline]{@Bayan, please add details, add prompts in the appendix }
% \textcolor{red}{HASNAT: Please add solid references of existing work that properly defines the term "silver labels"? If it does not exists in literature then we should make sure that this term is well defined by us with proper justification.}
% We additionally release a set $\sim$$67$K memes as an auxiliary
% resource, labeled by Gemini-3.1-Pro with a hatefulness label, subtype, propaganda, sentiment, and metadata.  these are not human-verified and support weakly-supervised use rather than as a core contribution (App.~\ref{app:silver}).

%% file: sections/experiments.tex
\section{Experiments}
\label{sec:experiments}

% We benchmark a broad range of model families on the two subtasks induced by our
% annotation scheme (\S\ref{ssec:annotation}). The first is binary hateful-meme
% detection (\emph{Hateful} vs.\ \emph{Not-Hateful}), and the second is fine-grained
% multi-label categorisation over the ten attack-strategy and pragmatic-function
% labels. Following the convention of related Arabic shared tasks, we refer to these
% as the \textbf{Binary} (A1) and \textbf{Subtype} (A2) tasks. Our goal is to
% establish reference performance under a leak-free protocol and to measure how far
% current models are from solving Arabic hateful-meme understanding.

We evaluate models on the two tasks defined by our annotation scheme. The first is binary hateful-meme detection, where each meme is classified as \emph{Hateful} or \emph{Not-Hateful}. The second is fine-grained multi-label hate-type classification.

% \subsection{Tasks and Evaluation Measures}
\subsection{Evaluation Measures}
\label{ssec:tasks}
For the binary task, we use macro-F1 as the primary evaluation measure and additionally report accuracy and recall. For the fine-grained hate-type task, we use macro-F1 averaged over all labels as the primary measure, with micro-F1 reported as a complementary measure to account for overall multi-label performance.

% For Binary we report macro-F1 (the official measure, averaged over
% the two classes), accuracy, the recall of the safety-critical \emph{Hateful}
% class. For subtype, we report macro-F1 (our primary measure, averaged over the ten labels and therefore
% dominated by the rare classes) together with the frequency-weighted micro-F1. The
% large gap between the two reflects the long-tailed subtype %distribution of
% Table~\ref{tab:split_finegrained}.

\subsection{Models}
\label{ssec:models}
We conduct experiments with five model groups covering unimodal encoders, multimodal fusion models, and vision-language models (VLMs). The text-only baselines use OCR text with AraBERTv2~\citep{antoun2020arabert}, MARBERTv2~\citep{abdul-mageed-etal-2021-arbert}, CAMeLBERT-mix~\citep{inoue-etal-2021-interplay}, XLM-R-base~\citep{conneau-etal-2020-unsupervised}, and mBERT~\citep{devlin-etal-2019-bert}. 
The image-only baselines use the meme image with ViT-B/16~\citep{DBLP:journals/corr/abs-2010-11929}, BEiT-B/16~\citep{bao2022beitbertpretrainingimage}, Swin-B~\citep{liu2021swintransformerhierarchicalvision}, ConvNeXt-V2-tiny~\citep{DBLP:journals/corr/abs-2301-00808}, and DINOv2-B~\citep{oquab2024dinov2learningrobustvisual}. 
Each unimodal encoder is followed by a linear classification head. The late-fusion models concatenate pooled representations from the two strongest text encoders and the two strongest vision encoders. The concatenated vector is passed to a two-layer MLP with GELU activation, dropout of 0.1, and hidden size 512.

% \textcolor{blue}{
We further benchmark both open- and closed-weight vision-language models (VLMs), including \texttt{Qwen3-VL-8B-Instruct} (Qwen3-VL-8B)~\cite{xu2025qwen3}, Fanar-2-Oryx-IVU~\cite{fanarteam2026fanar20arabicgenerative}, Phi-3.5-Vision~\cite{abdin2024phi3technicalreporthighly}, SmolVLM-Instruct, Qwen3-VL-8B-Thinking, InternVL3.5-8B, Gemini-2.5-Pro, Gemini-2.5-Flash~\cite{comanici2025gemini}, and GPT-5~\cite{singh2025openai}. Unless otherwise stated, all VLMs are evaluated in the zero-shot setting. In addition, we fine-tune Qwen3-VL-8B and evaluate its performance under both fine-tuning and few-shot in-context learning settings.
We conduct few-shot experiments only with Qwen3-VL-8B to examine the effect of few-shot examples under a controlled setting, rather than exhaustively compare prompting across all VLMs.

\subsection{Training and Model Selection}
\label{ssec:impl}

All fine-tuned encoders and fusion models are trained with AdamW using a weight decay of 0.01, a linear learning-rate schedule, and 6\% warmup. For binary hate detection, we use a softmax classification head with cross-entropy loss. For fine-grained hate-type prediction, we use a sigmoid output layer with binary cross-entropy loss to support multi-label classification. For each model family, we tune the learning rate, number of epochs, and batch size, and select the best checkpoint according to development macro-F1. The implementation details are provided in Appendix~\ref{ssec:compute}.

Open-weight VLMs (Qwen3-VL-8B and Qwen3-VL-2B) are fine-tuned with low-rank adaptation using LoRA with rank 16 and $\alpha{=}32$, applied to all linear layers while keeping the vision tower frozen. These models are trained for 3 epochs with a learning rate of $1{\times}10^{-4}$, a cosine schedule, and \texttt{bfloat16} precision. For all supervised settings, models are trained on \texttt{train}, selected on \texttt{dev} by macro-F1, and evaluated once on the held-out \texttt{test} split using the single best configuration. For fine-grained hate-type prediction, the sigmoid decision threshold is tuned on the \texttt{dev} set. Zero-shot models are evaluated directly on \texttt{test}. 
% All experiments use a fixed random seed of 42 and are run on NVIDIA H200 GPUs.

%% file: sections/results.tex
\section{Results and Discussion}
\label{sec:results}

% We first present the main benchmark results across all five model families
% (\S\ref{ssec:main}), then study the effect of prompting on the zero-shot closed
% models (\S\ref{ssec:prompting}), examine the dominant error modes
% (\S\ref{ssec:erroranalysis}), and finally test retrieval-augmented few-shot ICL as
% a training-free alternative (\S\ref{ssec:fewshot}).

\subsection{Results Across Model Families}
\label{ssec:main}

% Table~\ref{tab:main_results} reports the best configuration of every model on both
% subtasks. Six findings stand out.

In Table~\ref{tab:main_results}, we report performance across all experimental settings including random and majority-class baselines. 
% All trained and prompting-based models outperform these baselines, indicating that they capture task-relevant signals beyond label priors.

\begin{table*}[t]
\centering
\small
\setlength{\tabcolsep}{5pt}
\begin{tabular}{@{}llccccc@{}}
\toprule
& & \multicolumn{3}{c}{\textbf{Binary}} & \multicolumn{2}{c}{\textbf{Fine-grained} } \\
\cmidrule(lr){3-5}\cmidrule(lr){6-7}
\textbf{Family} & \textbf{Model} & macro-F1 & Acc & Rec\textsubscript{Hate} & macro-F1 & micro-F1 \\
\midrule
\multirow{2}{*}{Baseline}
 & Majority & 0.399 & 0.663 & 0.000 & 0.050 & 0.295 \\
 & Random   & 0.500 & 0.553 & 0.337 & 0.125 & 0.233 \\
\midrule
\multirow{5}{*}{Text (FT)}
 & MARBERTv2      & 0.709 & 0.743 & 0.596 & 0.263 & 0.390 \\
 & CAMeLBERT-mix  & 0.689 & 0.726 & 0.567 & 0.252 & 0.375 \\
 & AraBERTv2      & 0.678 & 0.706 & 0.611 & 0.295 & 0.411 \\
 & XLM-R-base     & 0.650 & 0.716 & 0.418 & 0.259 & 0.415 \\
 & mBERT          & 0.637 & 0.667 & 0.564 & 0.245 & 0.364 \\
\midrule
\multirow{5}{*}{Image (FT)}
 & ConvNeXt-V2-tiny & 0.657 & 0.700 & 0.513 & 0.246 & 0.418 \\
 & Swin-base        & 0.657 & 0.686 & 0.585 & 0.258 & 0.365 \\
 & BEiT-base        & 0.651 & 0.708 & 0.451 & 0.255 & 0.385 \\
 & ViT-base         & 0.645 & 0.708 & 0.427 & 0.268 & 0.424 \\
 & DINOv2-base      & 0.589 & 0.681 & 0.309 & 0.233 & 0.349 \\
\midrule
\multirow{4}{*}{Fusion (FT)}
 & MARBERTv2$+$ViT   & 0.730 & 0.757 & 0.650 & 0.274 & 0.428 \\
 & MARBERTv2$+$BEiT  & 0.724 & 0.750 & \textbf{0.656} & 0.318 & 0.444 \\
 & AraBERTv2$+$ViT   & 0.706 & 0.741 & 0.591 & 0.279 & 0.416 \\
 & AraBERTv2$+$BEiT  & 0.698 & 0.739 & 0.549 & 0.298 & 0.426 \\
\midrule
\multirow{6}{*}{\makecell[l]{Open VLM\\(zero-shot)}}
 & Qwen3-VL-8B-Instruct & 0.643 & 0.743 & 0.318 & 0.176 & 0.352 \\
 & Fanar-2-Oryx-IVU     & 0.535 & 0.700 & 0.154 & 0.115 & 0.288 \\
 & Phi-3.5-vision       & 0.486 & 0.682 & 0.095 & 0.044 & 0.159 \\
 & SmolVLM-Instruct     & 0.473 & 0.559 & 0.229 & 0.152 & 0.341 \\
 & Qwen3-VL-8B-Thinking & 0.469 & 0.676 & 0.077 & 0.191 & 0.382 \\
 & InternVL3.5-8B       & 0.458 & 0.676 & 0.062 & 0.164 & 0.369 \\
\midrule
\multirow{2}{*}{\makecell[l]{Open VLM\\(fine-tuned)}}
 & Qwen3-VL-8B & \textbf{0.768} & \textbf{0.797} & \textbf{0.656} & 0.334 & 0.475 \\
 & Qwen3-VL-2B & 0.739 & 0.782 & 0.558 & 0.284 & 0.437 \\
\midrule
\makecell[l]{Open VLM\\(few-shot)}
 & \makecell[l]{Qwen3-VL-8B} & 0.708 & 0.771 & 0.454 & 0.241 & 0.405 \\
 % $+$ RRF\\($K{=}5$, training-free
\midrule
\multirow{3}{*}{\makecell[l]{Closed VLM\\(zero-shot)}}
 & Gemini-2.5-pro   & 0.711 & 0.774 & 0.457 & \textbf{0.340} & 0.465 \\
 & GPT-5            & 0.628 & 0.741 & 0.282 & 0.301 & 0.464 \\
 & Gemini-3.5-flash & 0.499 & 0.694 & 0.104 & 0.271 & \textbf{0.486} \\
\bottomrule
\end{tabular}
\vspace{-0.2cm}
\caption{Results on the \benchmark{} test split. FT: Fine-tuned 
% Fine-tuned (FT) encoders, fusion, and the fine-tuned open VLMs use the dev-selected configuration; the zero-shot open and closed VLMs use the \textsc{default} prompt. The few-shot row is the best training-free configuration from \S\ref{ssec:fewshot} (RRF retrieval, $K{=}5$). 
Bold marks the best value per column. Rec\textsubscript{Hate} is recall of the \emph{Hateful} class, and fine-grained hate-type macro-F1 is averaged over all ten categories.}
\label{tab:main_results}
\vspace{-0.3cm}
\end{table*}
\noindent
\textbf{Fine-tuned VLMs are strongest for binary task.}
Qwen3-VL-8B with LoRA achieves the best binary macro-F1 of 0.768.
% and the best accuracy of 0.797. 
It outperforms the best late-fusion model, the best zero-shot closed VLM, and the strongest text encoder. Qwen3-VL-2B also performs well, reaching 0.739 macro-F1, which is close to the best fusion models. This suggests that supervised multimodal model is the most effective approach 
% for Arabic hateful-meme detection 
when labeled data are available.
% The fine-tuned Qwen3-VL-8B reaches the best binary macro-F1 (0.768), ahead of
% multimodal late fusion, the best zero-shot closed model (Gemini-2.5-pro), and the
% best text encoder (MARBERTv2). Moeover, when combining it with BEiT (Late fusion), it perfomed comparably with fine-tuned Qwen3-2B. 

% \paragraph{Zero-shot open VLMs lag behind commercial models and fine-tuning.}
\noindent
\textbf{Zero-shot VLMs show low hateful-class recall.}
Qwen3-VL-8B-Instruct is the strongest zero-shot open VLM, reaching 0.643 binary macro-F1, followed by Fanar-2-Oryx-IVU. The remaining open VLMs perform close to the majority baseline on binary and show very low \emph{Hateful} recall. The best zero-shot open VLM falls behind Gemini-2.5-pro by about 7 macro-F1 and the fine-tuned Qwen3-VL-8B by about 13 points. Fine-tuning substantially changes this pattern, improving Qwen3-VL-8B from 0.643 to 0.768 on binary and from 0.176 to 0.334 on fine-grained hate-type prediction.

% Among the six open-weight VLMs we evaluate zero-shot, Qwen3-VL-8B-Instruct is the
% strongest on Binary with macro-F1 of 0.643, followed by Fanar-2-Oryx-IVU; the rest stay near the majority baseline, and SmolVLM is poorly calibrated (near random). Even the
% best zero-shot open model trails the best closed model by about 7 points and the
% fine-tuned Qwen3-VL-8B by about 12. On Subtype the zero-shot open VLMs are weaker
% still. The decisive factor is adaptation: LoRA fine-tuning lifts Qwen3-VL-8B by
% about 12 points on Binary and roughly doubles its Subtype score.

% \paragraph{The fine-grained Subtype task is far from solved.}
% The best Subtype macro-F1 is only 0.340 (zero-shot Gemini-2.5-pro), with fine-tuned Qwen3-VL-8B and late fusion close behind. All five families fall in a narrow band, and the large gap between macro-F1 and micro-F1 shows that the rare hateful subtypes, rather than the frequent \emph{Humor} and \emph{Sarcasm}
% classes, are what hold macro-F1 down. Here the zero-shot closed model is the best
% system, slightly ahead of every fine-tuned model: fine-tuning on 3{,}500 memes is
% not enough to learn the long tail.

\noindent
\textbf{Fine-grained hate-type classification remains difficult.} 
The best fine-grained hate-type macro-F1 is 0.340, achieved by Gemini-2.5-pro, with fine-tuned Qwen3-VL-8B close behind at 0.334. All model groups remain below 0.35 macro-F1. The gap between macro-F1 and micro-F1 suggests that models perform better on frequent labels such as \emph{Humor} and \emph{Sarcasm} than on rare hate types. Fine-tuning improves the performance, however, it does not fully mitigate the effect of label imbalance.

% \paragraph{Multimodal fusion helps, but the text carries the signal.}
% Within the fine-tuned encoders the ordering is consistently fusion, then text,
% then image, on both subtasks. Late fusion improves over the best text encoder and
% raises \emph{Hateful} recall. Image-only models are the weakest family throughout,
% which indicates that for Arabic memes the embedded text is the dominant cue and
% the image adds a secondary, complementary signal.

\noindent
\textbf{Fusion helps, but text provides the strongest unimodal signal.} 
Late fusion improves over the best text-only encoder on binary and increases \emph{Hateful} recall, with MARBERTv2$+$BEiT matching the highest recall of 0.656. Image-only encoders are competitive, however, they are less reliable as standalone models. This pattern suggests that embedded Arabic text carries much of the task signal, while visual content provides complementary context that is most useful when combined with text.

% \paragraph{Arabic encoders beat multilingual ones.}
% Among text encoders, the Arabic-specific MARBERTv2 and AraBERTv2 lead the
% multilingual XLM-R and mBERT by several points on Binary, which shows the value of
% Arabic-centric pre-training for dialect-rich meme text.

\noindent
\textbf{Arabic-specific encoders provide strong text baselines.} 
Among text-only models, MARBERTv2 achieves the best binary macro-F1, while AraBERTv2 achieves the best fine-grained hate-type macro-F1. Both Arabic-specific encoders outperform mBERT and XLM-R-base on binary, highlighting the value of Arabic-centric pre-training for dialect-rich meme text.

% \paragraph{All models lag English benchmarks.}
% Even the strongest system (0.768 on Binary, 0.340 on Subtype) falls well below the
% scores reported on English hateful-meme benchmarks, despite using families that
% are competitive in English. This confirms that Arabic memes pose distinct
% challenges, such as right-to-left OCR, dialectal variation, and culturally specific
% symbols, and motivates dedicated Arabic resources such as \benchmark{}.

% \subsection{Effect of Prompting on Zero-shot LLMs}
% \label{ssec:prompting}

% We also compare three prompting strategies for the closed models (full results in
% Appendix Table~\ref{tab:cot}). Two patterns emerge. First, model scale matters more
% than the prompt: the ordering Gemini-2.5-pro, GPT-5, then Gemini-3.5-flash holds
% under every prompt, and the gap between models is much larger than any effect of
% the prompt. Second, chain-of-thought reasoning gives only small gains that depend
% on capacity: it barely moves Binary, helps the larger models on Subtype, and even
% hurts Gemini-3.5-flash when the reasoning is in Arabic, which suggests that low-capacity Arabic reasoning is itself a source of error.

\subsection{Effect of Prompting} % on Zero-shot Closed VLMs}
\label{ssec:prompting}
We compare \textit{direct prompting} with \textit{rationale-style prompting} in English and Arabic for closed VLMs. Full results are reported in Appendix Table~\ref{tab:cot}. Two trends are consistent. \textit{First}, model choice has a larger effect than prompt format. Gemini-2.5-pro performs best across all prompting settings, followed by GPT-5 and Gemini-3.5-flash. \textit{Second}, rationale-style prompting yields limited and model-dependent gains. It has little effect on binary performance, improves fine-grained hate-type prediction for the larger models, and hurts Gemini-3.5-flash when rationales are requested in Arabic. This suggests that \textit{Arabic rationale generation can introduce additional errors} for lower-capacity models.

\subsection{Retrieval-Augmented Few-Shot ICL}
\label{ssec:fewshot}

\begin{figure}[t]
\centering
\includegraphics[width=0.7\linewidth]{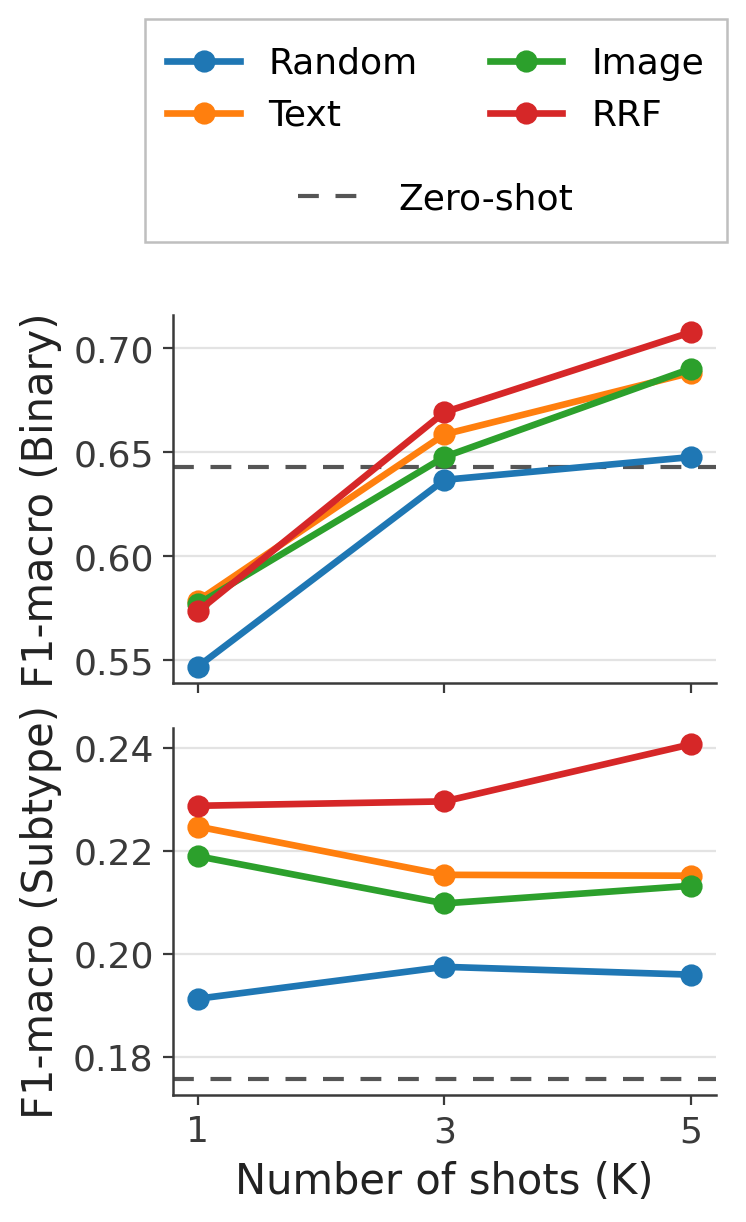}
\caption{Few-shot learning for Qwen3-VL-8B. Binary performance improves with more samples, however, it falls below zero-shot at $K{=}1$. Fine-grained hate-type performance improves modestly and saturates early. RRF is the strongest retrieval strategy.
}
\label{fig:fewshot}
\end{figure}

We evaluate retrieval-augmented few-shot ICL. For each test meme, we retrieve $K$ labeled examples from the \texttt{train} split and place them before the query as image-text demonstrations. We compare random demonstrations with text-based retrieval, image-based retrieval, and reciprocal rank fusion (RRF) over jina-clip-v2 embeddings~\citep{jina2024clipv2,cormack2009rrf}. We evaluate $K\in{1,3,5}$ using Qwen3-VL-8B and Qwen3-VL-2B. Details results  are reported in Appendix~\ref{sec:icl_appendix}.

\paragraph{Few-shot ICL gives moderate gains over zero-shot prompting, but remains below fine-tuning.} As shown in Figure~\ref{fig:fewshot}, RRF is the strongest retrieval strategy. The best setting is Qwen3-VL-8B with RRF and $K{=}5$, which improves binary macro-F1 from 0.643 to 0.708 and fine-grained hate-type macro-F1 from 0.176 to 0.241 over the zero-shot Qwen3-VL-8B baseline. However, this remains below full fine-tuning, which reaches 0.768 binary macro-F1 and 0.334 fine-grained hate-type macro-F1. These results show that retrieval selects more informative demonstrations but still falls short of supervised adaptation.
\paragraph{Retrieval mainly improves hateful-class recall.} RRF with $K{=}5$ raises \emph{Hateful} recall from 0.318 to 0.454, reducing the zero-shot model's bias toward non-hateful predictions. Gains for fine-grained hate-type prediction are smaller and are driven mostly by frequent labels such as \emph{Humor} and \emph{Mocking}, while rare hateful types remain difficult. A single demonstration is unstable and hurts binary performance, whereas $K{=}3$ and $K{=}5$ provide more reliable context.

\subsection{Error Analysis}
\label{ssec:erroranalysis}

\paragraph{Low recall is the main zero-shot failure.} 
Zero-shot VLMs show low \emph{Hateful} recall on binary. They under-predict the \emph{Hateful} class and therefore miss many hateful memes, making this the most important safety failure for moderation. Fine-tuning mitigates this problem, Qwen3-VL-8B and the best fusion model recover roughly two-thirds of hateful memes. This suggests that the low-recall pattern is mainly a limitation of the zero-shot setting rather than a limitation of the model family.

\noindent
\textbf{Fine-grained hate-type predictions are biased toward frequent labels.} 
The zero-shot closed VLMs predict fine-grained hate-type distributions that differ considerably from the gold label distribution, as shown in Appendix Table~\ref{tab:overpred}. Gemini-3.5-flash over-predicts frequent labels such as \emph{Humor} and \emph{Sarcasm}, which improves micro-F1, however,  weakens macro-F1. Gemini-2.5-pro over-predicts hateful types such as \emph{Mocking}, \emph{Contempt}, and \emph{Slurs}, increasing recall at the expense of precision. GPT-5 is more cautious. \emph{Exclusion} has only three test instances and is almost never predicted correctly.
% , further limiting macro-F1.

\noindent
\textbf{Longer meme text increases difficulty.} 
Binary errors increase with the length of the extracted Arabic text for all models, as shown in Appendix Table~\ref{tab:textlen}. The trend is steepest for Gemini-3.5-flash. Longer text often require more context, pragmatic inference, and target identification, whereas memes with short text are more likely to contain direct lexical cues. Larger models are more robust, however, the trend remains visible across systems.

\noindent
\textbf{Implicit hate is harder to classify.}
Qualitative inspection of Gemini-2.5-pro with Arabic rationale prompting shows that false negatives are not caused by OCR or image-understanding failures. The model identifies the visual template and reads dialectal Arabic correctly but treats implicit sectarian, religious, or misogynistic attacks as humor or satire and predicts \emph{Not-Hateful}. False positives are offensive dialectal jokes that annotators judged not to be hateful. These cases show that the hardest boundary is the normative judgment of implicit hate rather than multimodal prediction. This motivates the explicit taxonomy and annotation guidelines described in \S\ref{ssec:agreement}.

%% file: sections/conclusion.tex
\section{Conclusion and Future Work}
\label{sec:conclusions}

We presented \benchmark{}, a fine-grained Arabic hateful-meme benchmark comprising 5K human-annotated memes with a hierarchical annotation scheme covering hatefulness, attack subtype, and target category. We also release a separate auxiliary corpus of $\sim$66K silver-labeled memes with model-generated labels and metadata for weak supervision rather than evaluation. Experiments show that fine-tuning an open-weight VLM gives the best binary performance, while late fusion and Arabic text encoders remain strong baselines. Text provides the strongest unimodal signal, while visual context offers complementary information. Fine-grained classification remains challenging because models struggle to recover rare and overlapping hate categories. Zero-shot VLMs make conservative predictions and often miss hateful memes, while few-shot prompting provides only modest gains.
Future work will focus on rare and co-occurring hate categories, target-aware modeling, clearer separation between offensive humor and protected-group hate, and broader coverage of dialects and emerging meme formats.

%% file: sections/appendix.tex
\section{Data and Resource Release}
\label{sec:data_release}
We will release\footnote{\url{anonymous.com}} the supporting code, documentation, and dataset under the CC BY-NC-SA 4.0 license to facilitate adoption, reproducibility, and future extensions.

\section{Annotation Guidelines}
\label{sec:annotation_guidelines}

% This appendix reproduces the annotation guidelines provided to the annotators.
% We first present the English guidelines, including the definitions, the
% hierarchical task structure, and worked examples for every category
% (Appendix~\ref{ssec:guidelines_en}); we then provide the Arabic version that was
% used during annotation (Appendix~\ref{ssec:guidelines_ar}). The guidelines were
% authored in English, translated into Arabic, and reviewed by native
% Arabic-speaking NLP experts. \textcolor{red}

% {\textbf{Content warning:} the
% example memes in this appendix contain hateful and offensive material and are
% included solely for documentation.}

\subsection{English Annotation Guideline}
\label{ssec:guidelines_en}

\paragraph{Purpose.}
The purpose of this annotation is: \textit{(i)}~to identify whether a meme is \emph{Hateful} or \emph{Not-Hateful}; \textit{(ii)}~for hateful memes, to identify the attack type(s) (multi-label) and the targeted protected category(ies) (multi-label); and \textit{(iii)}~for non-hateful memes, to assign one or more non-hateful subtypes (multi-label). The annotation design is \emph{hierarchical} and \emph{conditional}: the fine-grained tasks are completed 
based on the binary label decision, discussed below.

% only under the corresponding branch of the binary decision (Figure/Table in Section~\ref{ssec:annotation}).

\paragraph{Definitions.}
We define \emph{hateful} content as a direct or indirect attack on people on the basis of a protected characteristic, including ethnicity, race, nationality, immigration status, religion, caste, sex, gender identity, sexual orientation, and disability or disease. An \emph{attack} includes violent or dehumanizing speech, such as comparing people to non-human entities, statements of inferiority, calls for exclusion or segregation, or mockery of a hate crime. We apply two boundary rules:

\begin{itemize}[leftmargin=*,nosep]
  \item Attacking groups that themselves promote or carry out hate, such as terrorist organisations, is \emph{not} considered hate speech.
  \item A meme that is insulting or offensive but does not target a protected category is labelled \emph{Not-Hateful} under this scheme.
\end{itemize}

\subsubsection{Task 1: Binary Classification (\texttt{is\_hateful}; Hateful vs.\ Not-Hateful)}
The task is to select exactly one of \emph{Hateful} or \emph{Not-Hateful}. The choice determines the subsequent annotation task: if \emph{Hateful}, complete Tasks~2 if \emph{Not-Hateful}, complete Task~3.

\subsubsection{Task 2: Attack Type (\texttt{hateful\_type}; Hateful only; multi-label)}
The task is to select  \emph{all} attack types that apply; a meme may exhibit several at once (e.g., \emph{Slurs}~$+$~\emph{Dehumanization}, or \emph{Mocking}~$+$~\emph{Inferiority}).

\begin{figure*}[t]
  \centering
  \includegraphics[trim={1cm 0cm 0cm 2cm}, clip, width=1\textwidth]{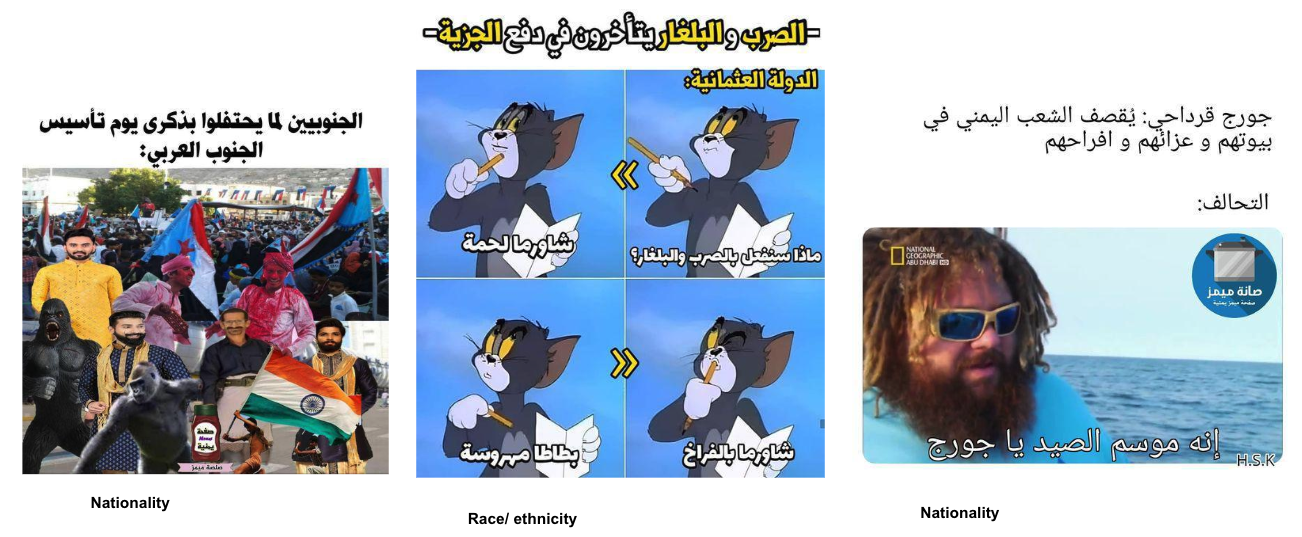}
  \caption{Example memes annotated as \emph{Hateful}~/ \emph{Dehumanization}.}
  \label{fig:ex_dehumanization}
\end{figure*}

\begin{figure*}[t]
  \centering
  \includegraphics[width=1\textwidth]{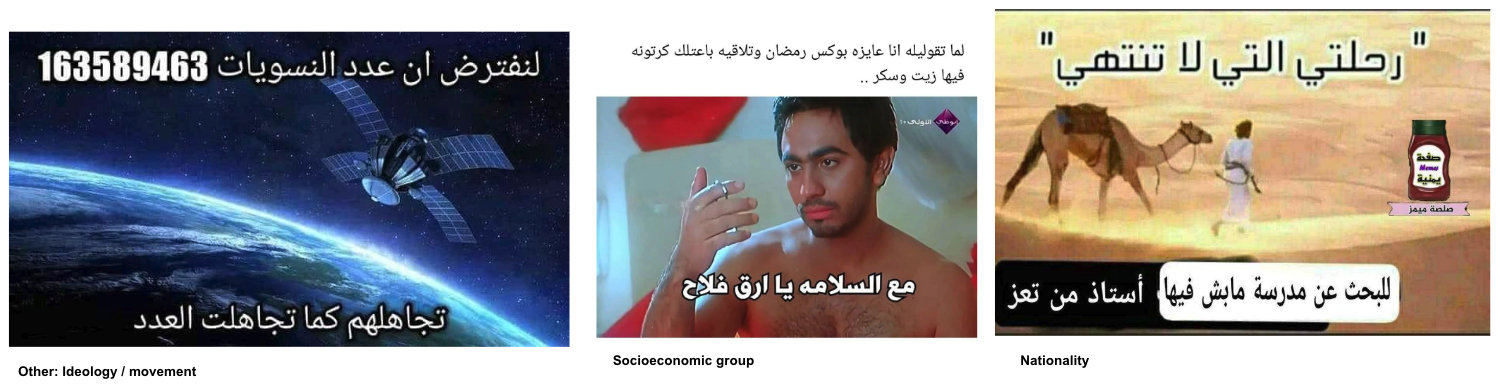}
  \caption{Example memes annotated as \emph{Hateful}~/ \emph{Inferiority}.}
  \label{fig:ex_inferiority}
\end{figure*}

\begin{figure*}[t]
  \centering
  \includegraphics[width=0.99\textwidth]{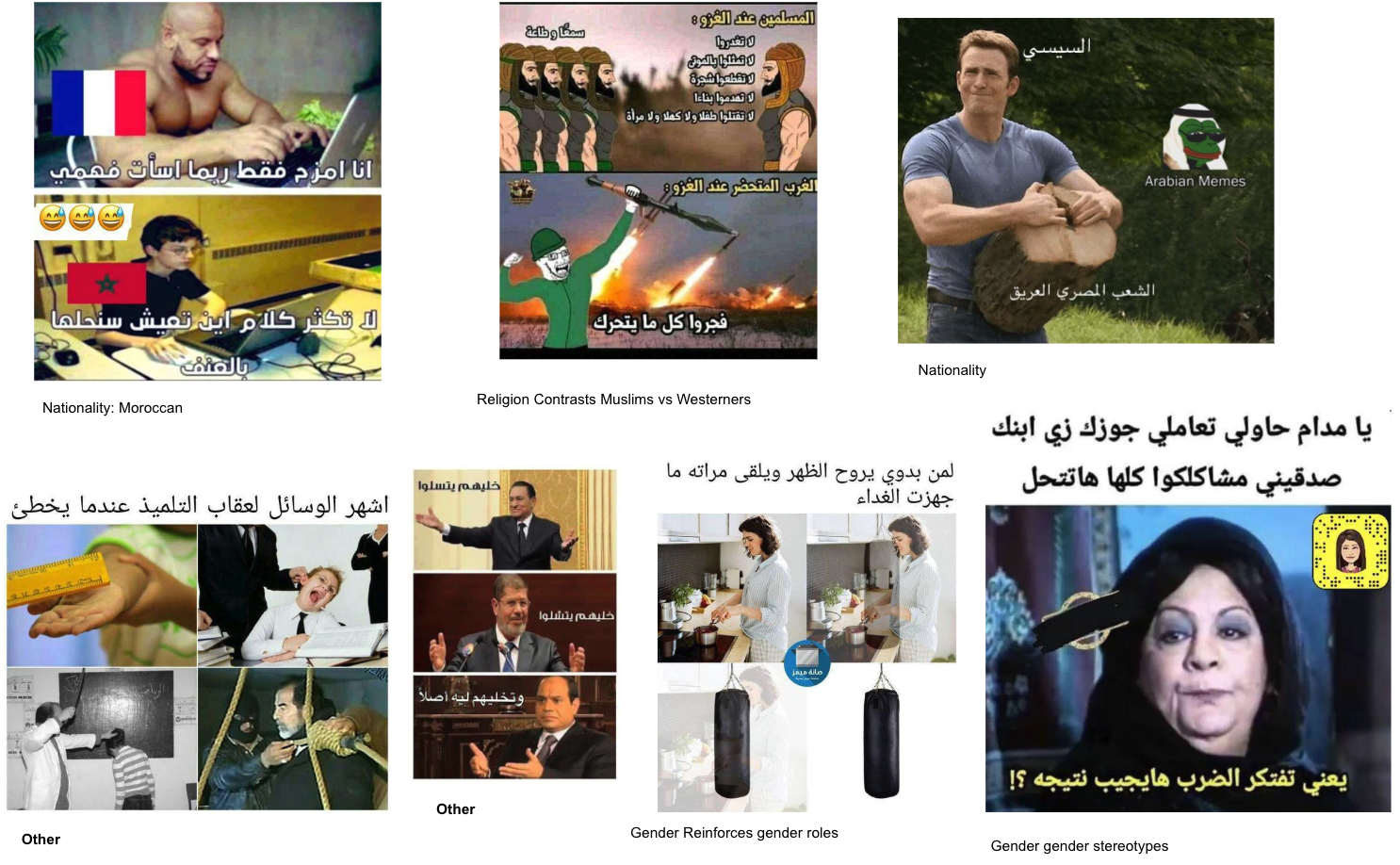}
  \caption{Example memes annotated as \emph{Hateful}~/ \emph{Incitement}.}
  \label{fig:ex_incitement}
  \vspace{-0.2cm}
\end{figure*}

\begin{figure*}[t]
  \centering
  \includegraphics[width=0.99\textwidth]{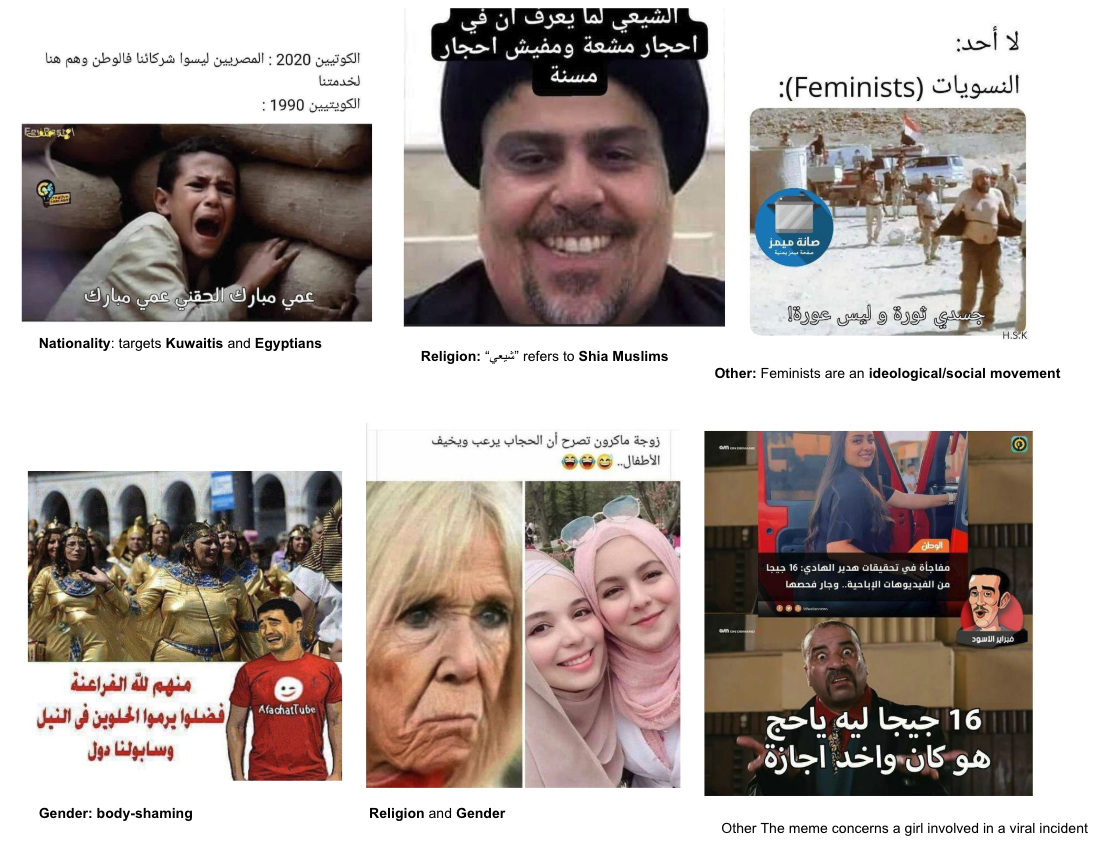}
  \caption{Example memes annotated as \emph{Hateful}~/ \emph{Mocking}.}
  \label{fig:ex_mocking}
  \vspace{-0.2cm}
\end{figure*}

\begin{figure*}[t]
  \centering
  \includegraphics[width=0.99\textwidth]{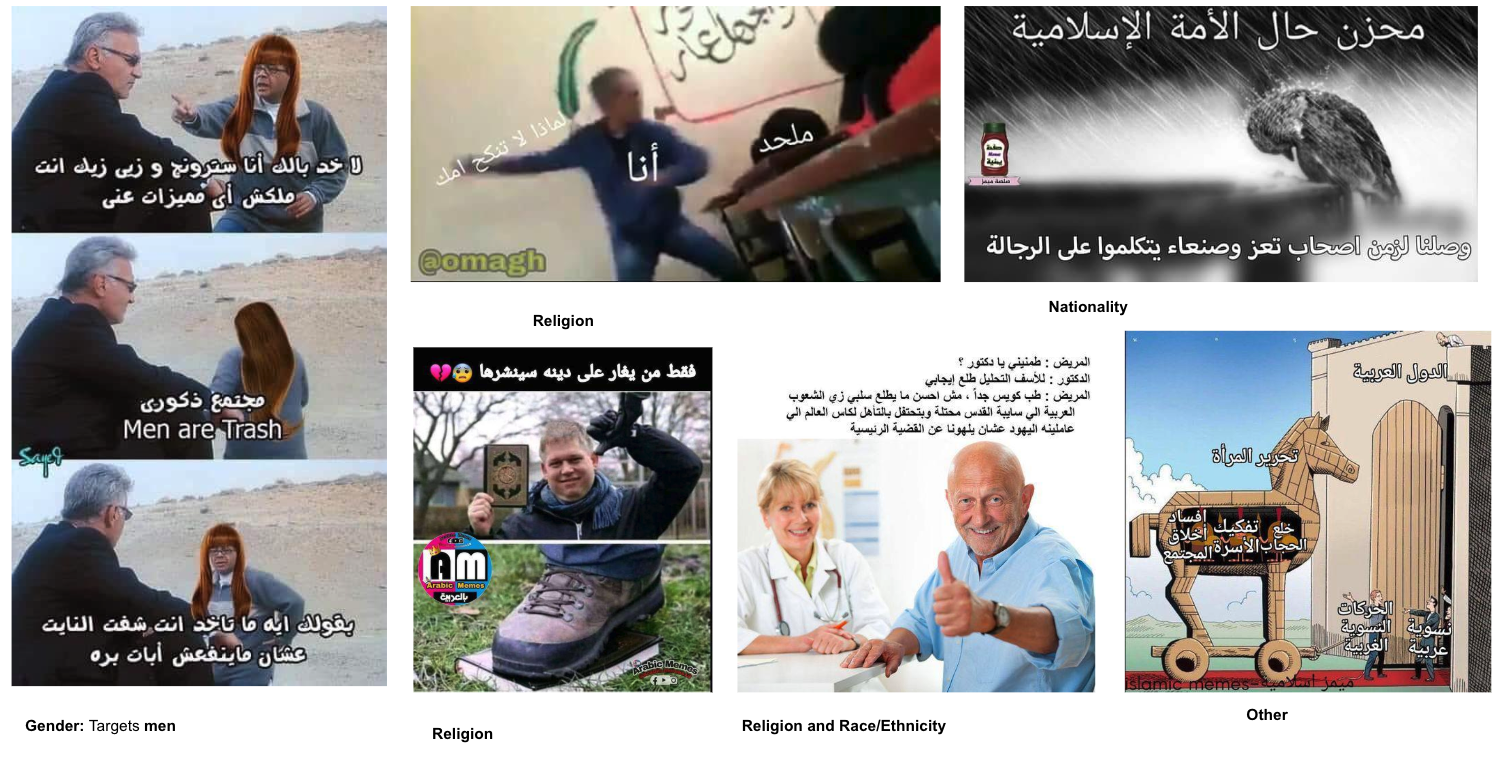}
  \caption{Example memes annotated as \emph{Hateful}~/ \emph{Contempt}.}
  \label{fig:ex_contempt}
  \vspace{-0.2cm}
\end{figure*}

\begin{figure*}[t]
  \centering
  \includegraphics[width=0.99\textwidth]{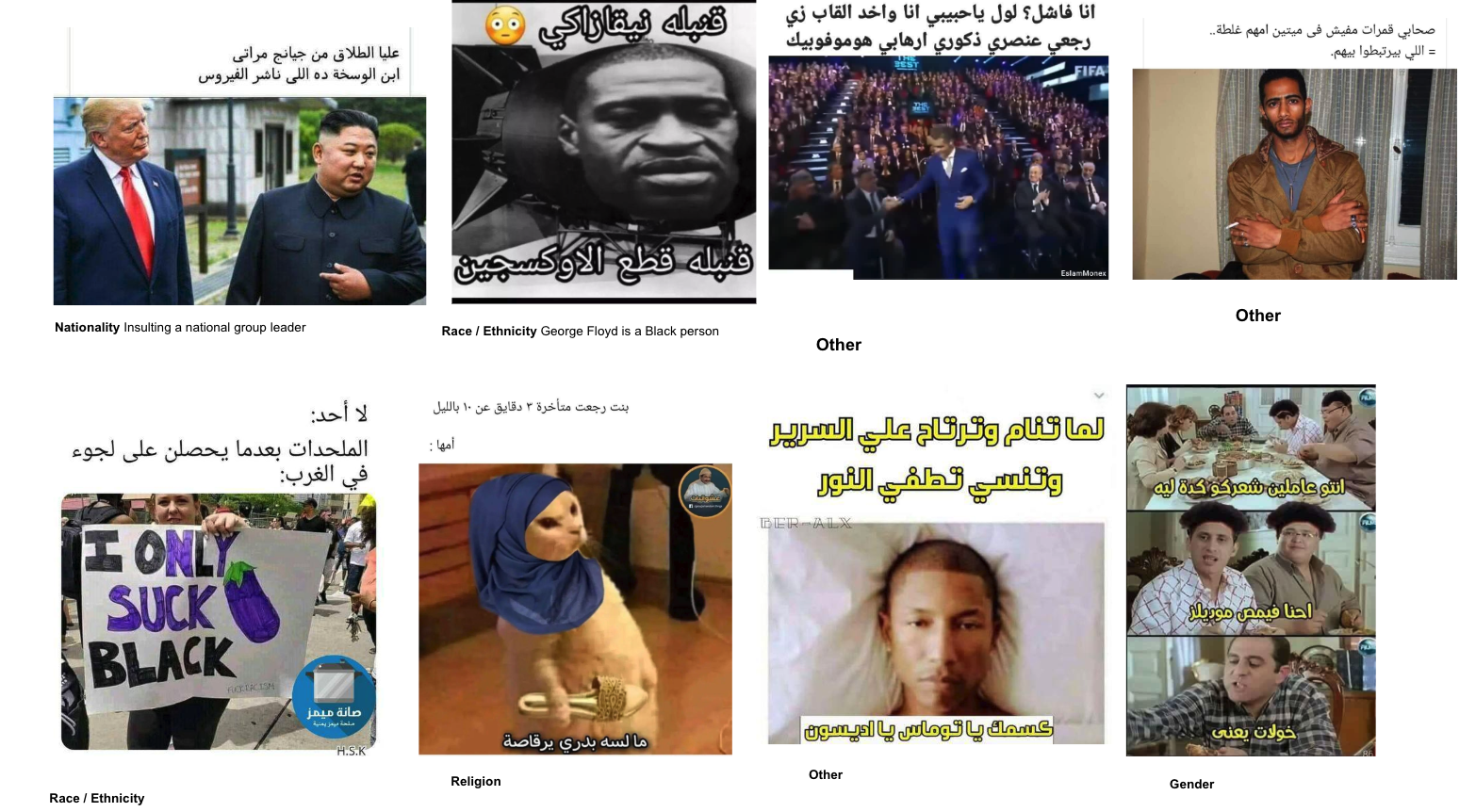}
  \caption{Example memes annotated as \emph{Hateful}~/ \emph{Slurs}.}
  \label{fig:ex_slurs}
  \vspace{-0.2cm}
\end{figure*}

\begin{figure*}[t]
  \centering
  \includegraphics[width=0.99\textwidth]{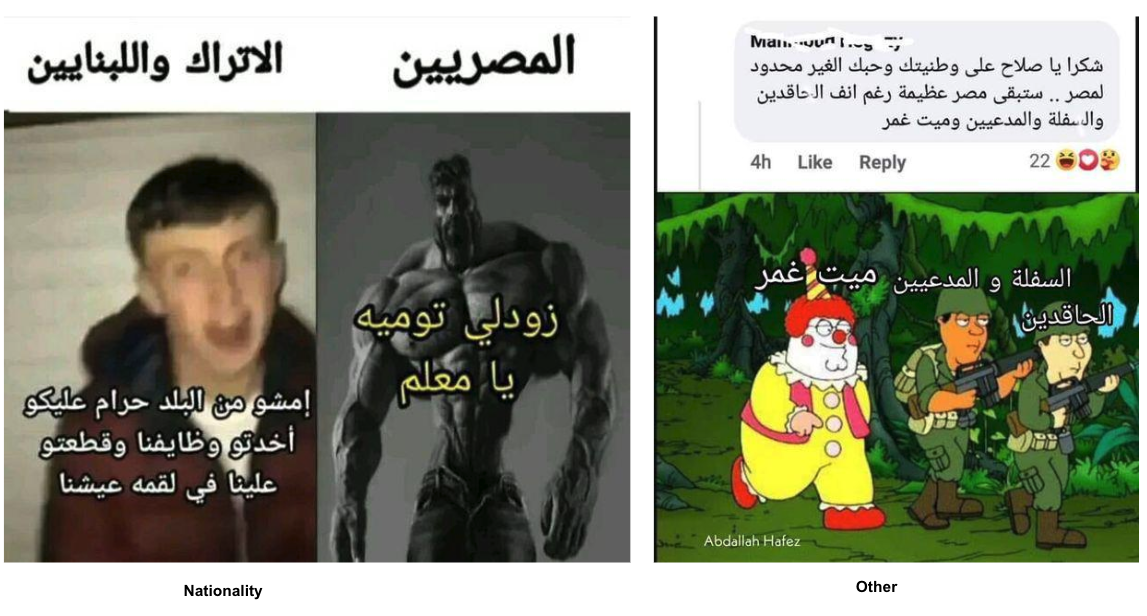}
  \caption{Example annotated as \emph{Hateful}~/ \emph{Exclusion}.}
  \label{fig:ex_exclusion}
  \vspace{-0.2cm}
\end{figure*}

\begin{itemize}[leftmargin=*,nosep]
\item \textbf{Dehumanization:} explicitly or implicitly describing a group as
subhuman, such as by comparing an ethnic group to animals or using ``fish
season'' metaphors (Figure~\ref{fig:ex_dehumanization}).

\item \textbf{Inferiority:} claiming that a group is inferior, less worthy, or
less important than society or another group, such as through class-based
belittling of ``peasants'' (Figure~\ref{fig:ex_inferiority}).

\item \textbf{Incitement:} explicitly or implicitly calling for harm to be
inflicted on a group, including physical attacks, such as through imagery of
forced physical discipline (Figure~\ref{fig:ex_incitement}).

\item \textbf{Mocking:} joking about, undermining, belittling, or disparaging
a group, such as by portraying its members as buffoonish or body-shaming a
gender (Figure~\ref{fig:ex_mocking}).

\item \textbf{Contempt:} expressing intensely negative feelings toward a
group, such as through statements like ``Men are trash'' or expressions of
ideological contempt (Figure~\ref{fig:ex_contempt}).

\item \textbf{Slurs:} using prejudicial terms to refer to or characterize a
group, including racial, religious, or gender-based slurs in Arabic dialects
(Figure~\ref{fig:ex_slurs}).

\item \textbf{Exclusion:} advocating, planning, or justifying the exclusion or
segregation of a group, such as through statements like ``Go back to your
country'' (Figure~\ref{fig:ex_exclusion}).

\item \textbf{Other:} a hateful attack that does not clearly fall under any of
the types above. This option is selected only when no other type applies and
cannot be combined with another attack type.
\end{itemize}

\subsubsection{Task 3: Not-Hateful Subtype (\texttt{non\_hateful\_type}; Not-Hateful only; multi-label)}
Non-hateful memes contain humorous, neutral, or positive content that does not target or harm a protected group, are intended for entertainment, and do not promote violence, hatred, or discrimination. The task is to select one or more subtypes.

\begin{figure*}[t]
  \centering
  \includegraphics[width=0.99\textwidth]{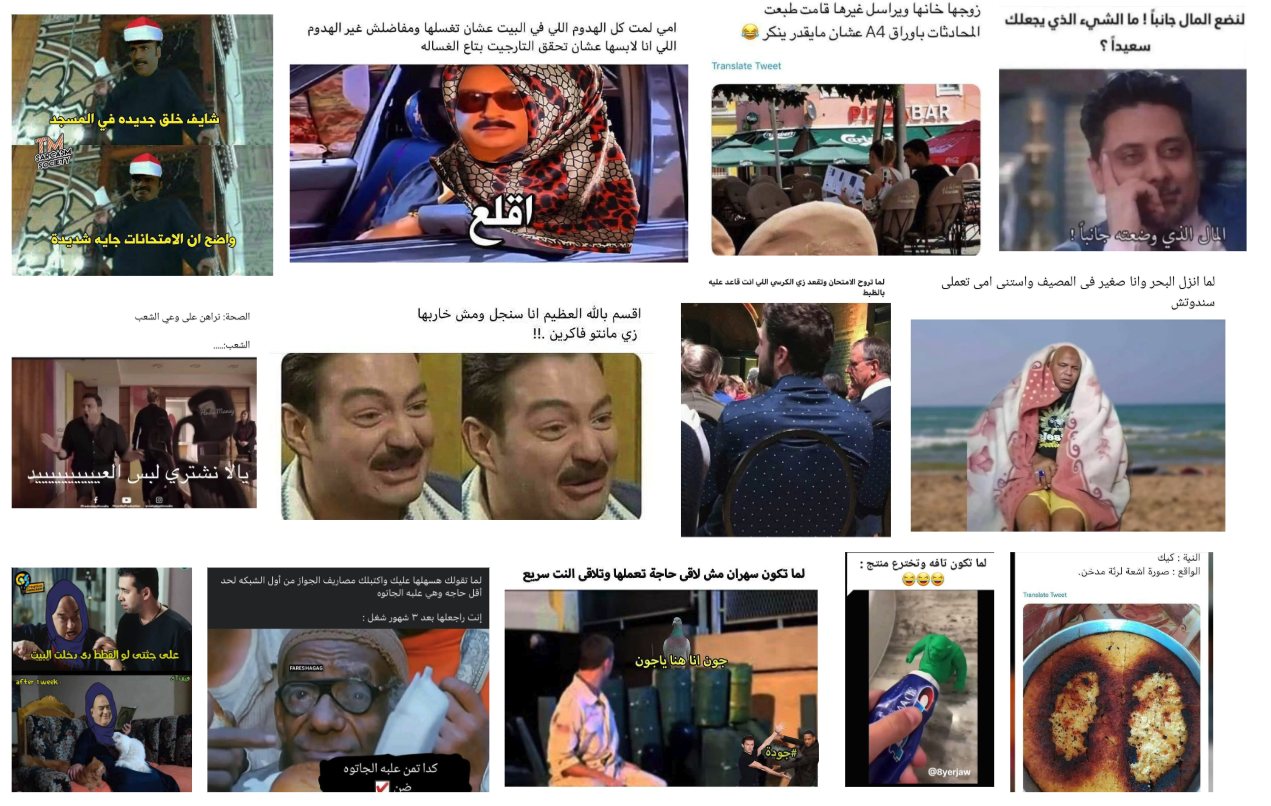}
  \caption{Example memes annotated as \emph{Not-Hateful}~/ \emph{Humor}.}
  \label{fig:ex_humor}
  \vspace{-0.2cm}
\end{figure*}

\begin{figure*}[t]
  \centering
  \includegraphics[width=0.99\textwidth]{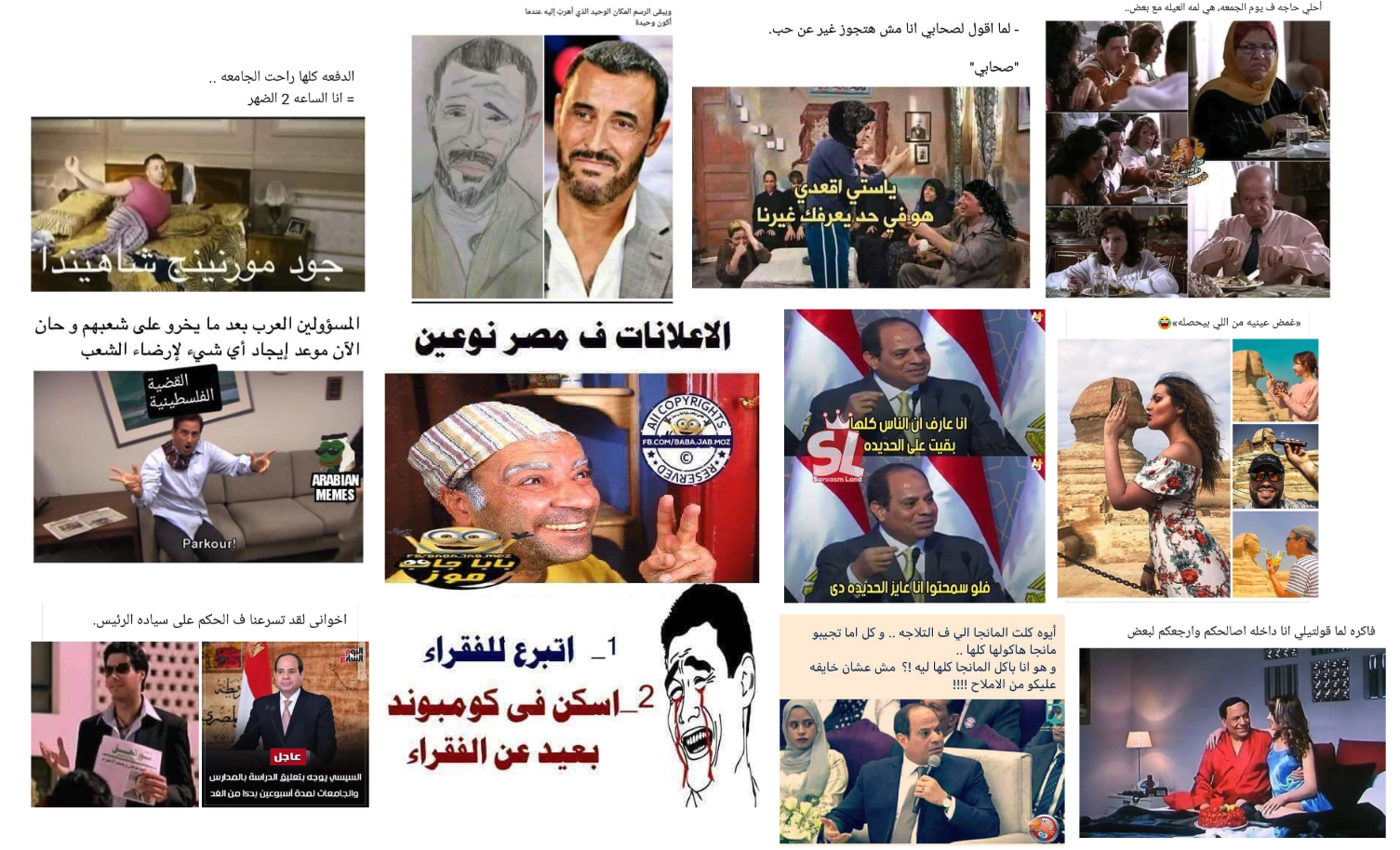}
  \caption{Example memes annotated as \emph{Not-Hateful}~/ \emph{Sarcasm}.}
  \label{fig:ex_sarcasm}
  \vspace{-0.2cm}
\end{figure*}

\begin{itemize}[leftmargin=*,nosep]
  \item \textbf{Humor:} content intended to entertain or amuse through jokes, puns, exaggeration, or playful commentary on everyday situations (work, family, food, travel), with no negative implication toward a protected group (Figure~\ref{fig:ex_humor}).
  \item \textbf{Sarcasm:} irony in which the intended meaning is the opposite of what is stated, used for humorous effect and directed at general behaviour or institutions (e.g., government, bureaucracy, inflation) rather than a protected group (Figure~\ref{fig:ex_sarcasm}).
  \item \textbf{Other:} neutral, positive, or informational memes that are not clearly humour or sarcasm (e.g., motivational quotes, announcements, generic reactions) and do not target a protected group.
\end{itemize}

% ------------------- Example figures (English guideline) -------------------

% =========================== ARABIC GUIDELINE ===========================
\subsection{Arabic Annotation Guideline}
\label{ssec:guidelines_ar}

The following is the Arabic version of the guidelines used during annotation.

\paragraph{Purpose and definitions.}
\noindent\includegraphics[width=\linewidth]{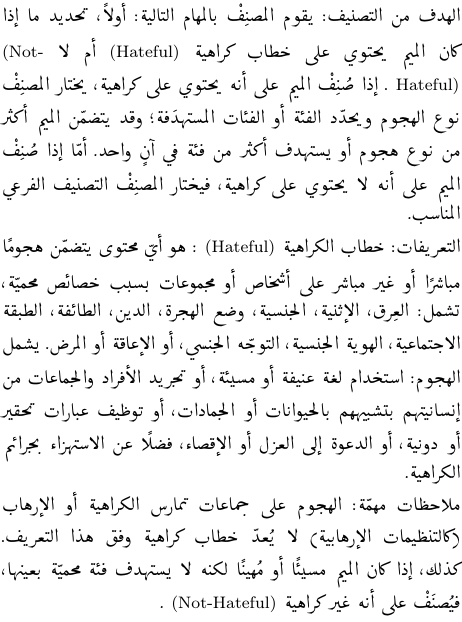}

\paragraph{Task 1: Meme classification.}
\noindent\includegraphics[width=\linewidth]{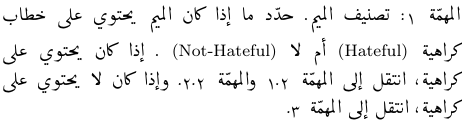}

\paragraph{Task 2: Attack types (multi-label).}
\noindent\includegraphics[width=\linewidth]{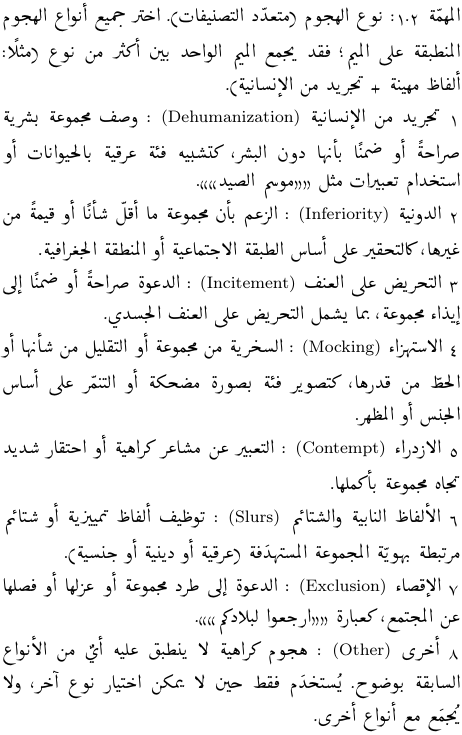}

\section{Experimental Details}
\label{sec:details_of_exp}

% This appendix complements Section~\ref{sec:experiments} with the prompt design,
% training and compute details, and additional dataset analysis. The dataset,
% guidelines, prompts, and evaluation scripts are released with the paper.

\subsection{Prompting the Zero-shot VLMs}
\label{ssec:prompts}
Each zero-shot query combines a \emph{system} message with an \emph{instruction} message. The system message frames the model as an expert Arabic multimodal content-moderation annotator covering MSA and the Egyptian, Levantine, Gulf, and Maghrebi dialects.. The instruction provides the meme image and OCR-extracted text and restricts the output to the predefined label space. For binary classification, the model predicts \emph{Hateful} or \emph{Not-Hateful} from the combined image--text content, guided by definitions that distinguish identity-based hate from general offensiveness. It returns the prediction as \texttt{{"label": \dots}}. For fine-grained classification, the prompt lists the ten taxonomy labels with brief definitions and asks for all applicable labels as a JSON list. The \textsc{cot-en} and \textsc{cot-ar} variants add a free-text \texttt{reasoning} field in English or Arabic, respectively. We fine-tune the open-weight VLMs with the same binary and fine-grained templates to generate the target label set directly.

\subsection{Training and Compute} 
\label{ssec:compute} 
We fine-tune the text, image, and fusion models with the HuggingFace \texttt{Transformers} \texttt{Trainer}. We use AdamW with weight decay 0.01, a linear learning-rate schedule, 6\% warmup, and 128-token text inputs. Hyperparameters are selected from the search space as reported in Table~\ref{tab:grid}, and the best checkpoint is chosen based on macro-F1 on the development set. For fine-grained hate-type prediction, we tune a global sigmoid threshold on the development set over $\{0.05,0.10,\dots,0.90\}$. Open-weight Qwen3-VL models are fine-tuned with \texttt{ms-swift} using LoRA with rank 16 and $\alpha{=}32$, applied to all linear layers while freezing the vision encoder. We train for 3 epochs with learning rate $1{\times}10^{-4}$, a cosine schedule, and \texttt{bfloat16}. All experiments use seed 42 on NVIDIA H200 GPUs. 
% Zero-shot closed models are evaluated using batch inference endpoints.

\begin{table}[t]
\centering
\small
\setlength{\tabcolsep}{3pt}
\begin{tabular}{@{}lcccc@{}}
\toprule
\textbf{Hyper-param.} & \textbf{Text} & \textbf{Image} & \textbf{Fusion} & \textbf{VLM} \\
\midrule
Learning rate & \makecell{2,3,5\\$\times10^{-5}$} & \makecell{3,5\\$\times10^{-5}$} & \makecell{2,3\\$\times10^{-5}$} & \makecell{1\\$\times10^{-4}$} \\
Epochs        & 5, 10 & 10, 15 & 5, 10 & 3 \\
Batch size    & 16, 32 & 32 & 16 & 4 \\
\bottomrule
\end{tabular}
\caption{Grid-search space for the fine-tuned families (best by dev macro-F1).
The open VLMs use fixed LoRA settings rather than a grid. The fusion-head learning
rate is fixed at $1{\times}10^{-4}$. All runs use seed 42.}
\label{tab:grid}
\vspace{-0.2cm}
\end{table}

\subsection{Additional Dataset Analysis}
\label{ssec:data_analysis}

\paragraph{OCR-text length.} In Table~\ref{tab:app_textlen}, we report the distribution
of OCR-extracted text length. The splits are well matched (mean
$\approx$\,86-88 characters), and \emph{Hateful} memes carry noticeably more text
than \emph{Not-Hateful} ones on the test split (96.2 vs.\ 83.1 characters),
consistent with the finding (Section~\ref{ssec:erroranalysis}) that hateful intent
is often conveyed through longer, more contextual captions.

\paragraph{Multi-label distribution.} In Table~\ref{tab:app_card}, we report the number of fine-grained categories assigned to each meme. The test split shows the greatest category overlap, with an average of 1.25 labels per meme; 21.4\% of test memes receive at least two categories, compared with 9.9\% in the training split.

% ,
% reflecting that the triple-annotated gold memes concentrated in the test set
% capture richer, co-occurring hate strategies.

\paragraph{Label co-occurrence.} Table~\ref{tab:app_cooc} shows the most frequent
fine-grained pairs. \emph{Mocking} acts as a hub, co-occurring with
\emph{Dehumanization}, \emph{Incitement}, \emph{Contempt}, and \emph{Slurs},
while among non-hateful memes \emph{Humor} and \emph{Sarcasm} frequently appear
together. These patterns motivate modelling hate as co-occurring strategies
rather than a single label per meme.

\begin{table}[t]
\centering
% \small
\setlength{\tabcolsep}{2pt} 
\scalebox{0.85}{
\begin{tabular}{@{}lrrr@{}}
\toprule
\textbf{Split} & \textbf{Mean} & \textbf{Median} & \textbf{Max} \\
\midrule
Train & 85.5 & 76 & 641 \\
Dev   & 87.8 & 80 & 619 \\
Test  & 87.5 & 77 & 644 \\
% \midrule
% \multicolumn{4}{@{}l}{\emph{Test by class:} Hateful 96.2 \quad Not-Hateful 83.1 (mean)} \\
\bottomrule
\end{tabular}
}
\caption{OCR-text length (characters) per split. \emph{Test set class-wise:} Hateful 96.2; Not-Hateful 83.1 (mean)}
\label{tab:app_textlen}
\vspace{-0.2cm}
\end{table}

% \begin{table}[t]
% \centering
% \small
% \begin{tabular}{@{}lrrrr@{}}
% \toprule
% \textbf{Split} & \textbf{Mean} & \textbf{Median} & \textbf{P90} & \textbf{Max} \\
% \midrule
% Train & 85.5 & 76 & 139 & 641 \\
% Dev   & 87.8 & 80 & 138 & 619 \\
% Test  & 87.5 & 77 & 138 & 644 \\
% \midrule
% \multicolumn{5}{@{}l}{\emph{Test by class:} Hateful 96.2 \quad Not-Hateful 83.1 (mean)} \\
% \bottomrule
% \end{tabular}
% \caption{OCR-text length (characters) per split.}
% \label{tab:app_textlen}
% \end{table}

\begin{table}[t]
\centering
\small
\begin{tabular}{@{}lrrrrr@{}}
\toprule
% \textbf{Split} & \textbf{Avg} & \textbf{1} & \textbf{2} & \textbf{3} & \textbf{4} \\
\multicolumn{1}{l}{\textbf{Split}} &
\multicolumn{1}{c}{\textbf{Avg}} &
\multicolumn{1}{c}{\textbf{1}} &
\multicolumn{1}{c}{\textbf{2}} &
\multicolumn{1}{c}{\textbf{3}} &
\multicolumn{1}{c}{\textbf{4}} \\
\midrule
Train & 1.11 & 3{,}154 & 298 & 48 & 0 \\
Dev   & 1.15 & 439 & 47 & 14 & 0 \\
Test  & 1.25 & 786 & 178 & 32 & 4 \\
\bottomrule
\end{tabular}
\caption{Fine-grained label cardinality: number of memes with 1-4 fine-grained labels, and
the mean number of labels per meme.}
\label{tab:app_card}
\vspace{-0.2cm}
\end{table}

\begin{table}[t]
\centering
\small
\begin{tabular}{@{}lr@{}}
\toprule
\textbf{Fine-grained pair} & \textbf{Count} \\
\midrule
Dehumanization $+$ Mocking    & 145 \\
Incitement $+$ Mocking        & 114 \\
Humor $+$ Sarcasm             & 81 \\
Contempt $+$ Mocking          & 78 \\
Dehumanization $+$ Incitement & 66 \\
Mocking $+$ Slurs             & 60 \\
Inferiority $+$ Mocking       & 55 \\
Incitement $+$ Slurs          & 44 \\
Contempt $+$ Incitement       & 39 \\
Dehumanization $+$ Slurs      & 31 \\
\bottomrule
\end{tabular}
\vspace{-0.2cm}
\caption{Most frequent fine-grained hate-type co-occurrence pairs (all splits).}
\label{tab:app_cooc}
\vspace{-0.2cm}
\end{table}

\subsection{Effect of Prompting on the Closed Models}
\label{ssec:prompt_appendix}
Table~\ref{tab:cot} reports the full prompting comparison summarised in \S\ref{ssec:prompting}. For each closed model it gives macro-F1 under the direct prompt and under chain-of-thought reasoning in English and in Arabic. The model ordering is stable across prompts, and chain-of-thought helps the larger models on
fine-grained categories while hurting the small Gemini-3.5-flash when the reasoning is in Arabic.

\begin{table}[t]
\centering
\small
\setlength{\tabcolsep}{5pt}
\begin{tabular}{@{}llcc@{}}
\toprule
\textbf{Model} & \textbf{Prompt} & \textbf{Binary} & \textbf{Fine-grained} \\
\midrule
\multirow{3}{*}{Gemini-2.5-pro}
 & \textsc{default} & 0.711 & 0.340 \\
 & \textsc{cot-en}  & 0.709 & \textbf{0.358} \\
 & \textsc{cot-ar}  & \textbf{0.726} & 0.333 \\
\midrule
\multirow{3}{*}{GPT-5}
 & \textsc{default} & 0.628 & 0.301 \\
 & \textsc{cot-en}  & 0.642 & 0.313 \\
 & \textsc{cot-ar}  & 0.647 & 0.315 \\
\midrule
\multirow{3}{*}{Gemini-3.5-flash}
 & \textsc{default} & 0.499 & 0.271 \\
 & \textsc{cot-en}  & 0.499 & 0.272 \\
 & \textsc{cot-ar}  & 0.493 & 0.265 \\
\bottomrule
\end{tabular}
\vspace{-0.2cm}
\caption{Macro-F1 of the zero-shot closed VLMs under direct (\textsc{default}) and
chain-of-thought (English or Arabic) prompting. The best value per column is in
bold.}
\label{tab:cot}
\vspace{-0.2cm}
\end{table}

% \subsection{Additional Error Analysis}
% \label{ssec:error_appendix}
% This subsection holds the two supporting tables for the zero-shot error analysis
% of \S\ref{ssec:erroranalysis}. Table~\ref{tab:overpred} reports, for each label,
% how many memes each closed model predicts compared with the gold support, which
% shows the over-prediction of frequent tone labels and of hateful subtypes.
% Table~\ref{tab:textlen} reports the Binary error rate in three buckets of
% OCR-text length, which shows that longer captions are harder for all models and
% most of all for the smallest one.

\subsection{Additional Error Analysis}
\label{ssec:error_appendix}

We provide two supporting tables for the zero-shot error analysis in \S\ref{ssec:erroranalysis}. Table~\ref{tab:overpred} compares, for each label, the predicted count from each closed VLM with the corresponding gold label count. This highlights the over-prediction of frequent non-hateful labels such as \emph{Humor} and \emph{Sarcasm}, as well as several hateful fine-grained categories. Table~\ref{tab:textlen} reports binary error rates across three OCR-text length groups, showing that longer captions are harder for all models, with the largest degradation observed for the smallest model.

\begin{table}[t]
\centering
\small
\setlength{\tabcolsep}{5pt}
\begin{tabular}{@{}lrrrr@{}}
\toprule
\textbf{Label} & \textbf{Gold} & \textbf{Pro} & \textbf{GPT-5} & \textbf{Flash} \\
\midrule
Humor    & 332 & 728 & 878 & 885 \\
Sarcasm  & 333 & 349 & 404 & 869 \\
Mocking  & 211 & 465 & 164 & 120 \\
Contempt & 50  & 264 & 64  & 45  \\
Slurs    & 47  & 166 & 31  & 9   \\
\bottomrule
\end{tabular}
\vspace{-0.2cm}
\caption{Subtype label over- and under-prediction (predicted count vs.\ gold
support) for the zero-shot closed models under the \textsc{default} prompt.}
\label{tab:overpred}
\vspace{-0.2cm}
\end{table}

\begin{table}[t]
\centering
\small
\setlength{\tabcolsep}{5pt}
\begin{tabular}{@{}lccc@{}}
\toprule
\textbf{Model (\textsc{default})} & \textbf{21--60} & \textbf{61--120} & \textbf{120+} \\
\midrule
Gemini-2.5-pro   & 0.222 & 0.221 & 0.255 \\
GPT-5            & 0.239 & 0.260 & 0.299 \\
Gemini-3.5-flash & 0.258 & 0.312 & 0.382 \\
\bottomrule
\end{tabular}
\vspace{-0.2cm}
\caption{Binary error rate by OCR-text length (characters). Longer, more
contextual captions are harder, and the smallest model degrades most.}
\label{tab:textlen}
\vspace{-0.2cm}
\end{table}

% \section{Retrieval-Augmented In-Context Learning: Setup, Ablations, and Analysis}

\section{In-Context Learning}
\label{sec:icl_appendix}

\subsection{Few-Shot Retrieval Setup}
\label{ssec:icl_setup}
For each test meme, we retrieve demonstrations from the 3{,}500-meme training pool using a CLIP-style dual encoder. Our main retriever is jina-clip-v2~\citep{jina2024clipv2}, which embeds images and text into a shared multilingual space; we use SigLIP-2~\citep{zhai2023siglip} as an ablation. We construct each prompt as a multi-turn chat with a task system prompt, $K$ demonstration turns, and a final query turn. Each demonstration includes the meme image, its OCR text, and its gold label.

We compare text-only retrieval, image-only retrieval, and reciprocal rank fusion (RRF). Text and image retrieval rank candidates by cosine similarity in the corresponding modality. RRF combines the two rank lists as $s(d)=\sum_{m\in{t,v}}\frac{1}{60+\mathrm{rank}_m(d)}$~\citep{cormack2009rrf}, making the fusion independent of modality-specific score scales. 
% Embeddings are stored without normalisation to support the cosine-versus-Euclidean ablation.

Retrieval and prompt construction are performed using only the training dataset. 
% Across all configurations, we verify that no test item is used as a demonstration, no query retrieves itself, and all runs produce exactly 1{,}000 parsed predictions. Unless otherwise stated, 
We use RRF with $K{=}3$, cosine similarity, jina-clip-v2 embeddings, and similar-last ordering.
% , and the original unbalanced training pool.

% \subsection{Full Strategy $\times$ $K$ Grid}
% \label{ssec:icl_grid}
% Table~\ref{tab:app_fewshot8b} reports the complete grid for Qwen3-VL-8B and
% Table~\ref{tab:app_fewshot2b} for Qwen3-VL-2B. On the 8B, RRF is best on both
% subtasks and improves steadily up to $K{=}5$, and on Binary every strategy falls
% below zero-shot at $K{=}1$. The 2B instead follows an inverted-U pattern, where its
% score rises up to $K{=}3$ and then falls at $K{=}5$, and for this model image-only
% retrieval is as strong as RRF. The smaller model cannot make as much use of longer
% or fused demonstration contexts as the 8B.

% \subsection{Strategy and Demonstration Setup}
\subsection{Few-Shot Results}
\label{ssec:icl_grid}

In Tables~\ref{tab:app_fewshot8b} and~\ref{tab:app_fewshot2b}, we report the strategy-by-$K$ results for Qwen3-VL-8B and Qwen3-VL-2B, respectively. For Qwen3-VL-8B, RRF achieves the best performance on both subtasks and improves consistently up to $K{=}5$. On binary task, however, all strategies perform below zero-shot at $K{=}1$. Qwen3-VL-2B follows a different pattern, peaking at $K{=}3$ and dropping at $K{=}5$, while image-only retrieval matches RRF. These results suggest that the smaller model benefits from a small number of demonstrations, although it uses longer or fused demonstration contexts less effectively.

\begin{table}[t]
\centering
\setlength{\tabcolsep}{4pt}
\resizebox{\columnwidth}{!}{%
\begin{tabular}{@{}lcccc c ccc@{}}
\toprule
 & \multicolumn{4}{c}{\textbf{Binary} (macro-F1)} & & \multicolumn{3}{c}{\textbf{Fine-grained} (macro-F1)}\\
\cmidrule(lr){2-5}\cmidrule(lr){7-9}
Strategy & $K0$ & $K1$ & $K3$ & $K5$ & & $K1$ & $K3$ & $K5$\\
\midrule
zero-shot & 0.643 & --- & --- & --- & & --- & --- & ---\\
random & --- & 0.547 & 0.637 & 0.648 & & 0.191 & 0.197 & 0.196\\
text & --- & 0.579 & 0.659 & 0.688 & & 0.225 & 0.215 & 0.215\\
image & --- & 0.577 & 0.648 & 0.690 & & 0.219 & 0.210 & 0.213\\
RRF & --- & 0.574 & 0.669 & \textbf{0.708} & & 0.229 & 0.230 & \textbf{0.241}\\
\bottomrule
\end{tabular}%
}
\caption{Retrieval-augmented few-shot ICL on \textbf{Qwen3-VL-8B} (Fine-grained
zero-shot $=0.176$). RRF beats random and, at $K{\ge}3$, zero-shot. Best per task
in bold.}
\label{tab:app_fewshot8b}
\end{table}

\begin{table}[t]
\centering
\setlength{\tabcolsep}{4pt}
\resizebox{\columnwidth}{!}{%
\begin{tabular}{@{}lccc c ccc@{}}
\toprule
 & \multicolumn{3}{c}{\textbf{Binary}} & & \multicolumn{3}{c}{\textbf{Fine-grained}}\\
\cmidrule(lr){2-4}\cmidrule(lr){6-8}
Strategy & $K1$ & $K3$ & $K5$ & & $K1$ & $K3$ & $K5$\\
\midrule
random & 0.590 & 0.623 & 0.615 & & 0.131 & 0.140 & 0.145\\
text   & 0.639 & 0.658 & 0.629 & & 0.175 & 0.161 & 0.189\\
image  & 0.661 & 0.681 & 0.665 & & 0.160 & \textbf{0.210} & 0.174\\
RRF    & 0.663 & \textbf{0.684} & 0.671 & & 0.203 & 0.181 & 0.209\\
\bottomrule
\end{tabular}%
}
\vspace{-0.2cm}
\caption{Retrieval-augmented few-shot ICL on \textbf{Qwen3-VL-2B}. The small model follows an inverted-U pattern, rising up to $K{=}3$ and then falling at $K{=}5$, unlike the 8B.}
\label{tab:app_fewshot2b}
\vspace{-0.2cm}
\end{table}

% \paragraph{Model scale.} Figure~\ref{fig:icl_modelsize} plots the best F1 (over
% $K\in\{1,3,5\}$) per strategy for the two model sizes. The 8B is better than the 2B
% at every strategy on both subtasks, and the margin grows as the retrieval strategy
% improves. Moving from random to RRF lifts the 8B from 0.650 to 0.708 on Binary and
% from 0.203 to 0.241 on Subtype, a steeper climb than for the 2B. RRF is the best
% strategy for each model, but only the 8B turns the extra fused signal into steady
% gains, so few-shot ICL rewards the larger model more.

\begin{figure}[t]
\centering
\includegraphics[width=1\linewidth]{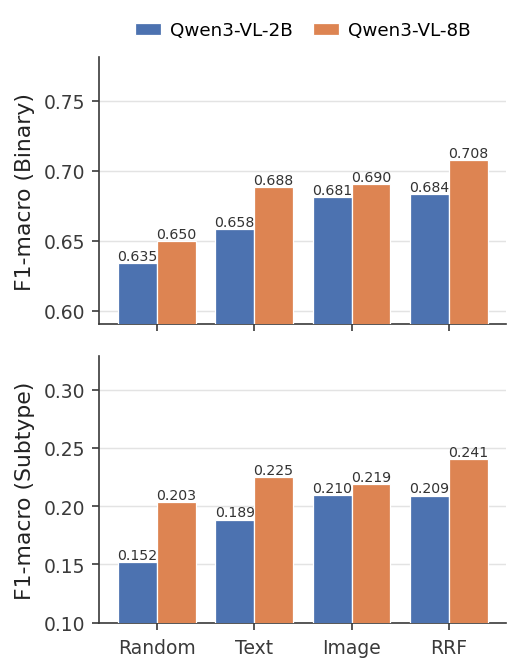}
\caption{Best F1 (max over $K\in\{1,3,5\}$) per retrieval strategy for Qwen3-VL-2B
vs.\ 8B. The 8B wins at every strategy; the gain from random to RRF is larger for
the 8B, on both Binary (top) and Subtype (bottom).}
\label{fig:icl_modelsize}
\end{figure}

\paragraph{Model scale.} 
In Figure~\ref{fig:icl_modelsize}, we compare the best F1 over $K\in\{1,3,5\}$ for each retrieval strategy and model size. Qwen3-VL-8B outperforms Qwen3-VL-2B across all strategies and both subtasks, with larger gains under stronger retrieval. Moving from random retrieval to RRF improves Qwen3-VL-8B from 0.650 to 0.708 on binary and from 0.203 to 0.241 on fine-grained hate-type prediction, while Qwen3-VL-2B gains less. RRF performs best for both models, however, only the 8B model consistently benefits from the fused retrieval signal. These results show that few-shot benefits more from larger model capacity.

\section{Details of Silver Dataset}
\label{app:silver}

% \paragraph{Selection and silver set.} After filtering (\S\ref{ssec:filtering}) the pool
% holds $\sim$$71$K text-bearing memes. Gemma-3-12B labelled all of them with a binary
% hateful/not-hateful prompt; the flagged memes were combined with randomly sampled
% non-flagged memes to form the $5$K annotation set (selector labels hidden from annotators
% and never used as gold). The $\sim$$66$K unselected memes are released with Gemini-3.1-Pro
% silver labels: of $66{,}297$ labelled memes ($116$ dropped for parse/label errors), only
% $580$ ($0.87\%$) are \emph{Hateful}, confirming the rarity that motivates model-assisted
% selection. For the $5$K gold memes we generate the same metadata \emph{conditioned} on the
% human label and subtype, so it never relabels them. Both prompts are listed in
% App.~\ref{app:prompts}.

\paragraph{Selection and silver labels.} 
After removing duplicates and filtering out memes without extracted text, we retain $\sim$71K memes. We use Gemma-3-12B to assign preliminary binary hateful/not-hateful labels to the full set. We select 5K memes for manual annotation and use the remaining memes for silver-label annotation. Prompts are listed in Section \S~\ref{app:prompts}.

% \paragraph{Silver-label distributions.}
% \label{ssec:silver_stats}
% In Tables~\ref{tab:silver_single} and~\ref{tab:silver_multi}, we summarise the Gemini-3.1-Pro silver labels over the $\sim66{,}297$ auxiliary memes. Beyond the $0.87\%$ hateful rate, the collection is strongly
% Egyptian and MSA, dominated by everyday and entertainment topics and a humorous register;
% among the targets that \emph{are} mentioned, \emph{Gender/Sex} is by far the most frequent,
% and propaganda appears in only $2.9\%$ of memes.

\paragraph{Silver-label distributions.}
\label{ssec:silver_stats}
In Tables~\ref{tab:silver_single} and~\ref{tab:silver_multi}, we summarize the Gemini-3.1-Pro silver labels for the $\sim$66K auxiliary memes. The silver set has a low hateful rate of 0.87\%. It is dominated by Egyptian Arabic and MSA, everyday and entertainment topics, and humorous content. Among memes that mention a target, \emph{Gender/Sex} is the most frequent target category. Propaganda is rare, appearing in only 2.9\% of memes.

\begin{table*}[t]
\centering
% \small
% \setlength{\tabcolsep}{3pt}
% \renewcommand{\arraystretch}{1.25}
\setlength{\tabcolsep}{2pt} 
\scalebox{0.85}{
\begin{tabular*}{\textwidth}{@{\extracolsep{\fill}} l r l r l r @{}}
\toprule
\multicolumn{6}{c}{\cellcolor{gray!12}\textbf{\textsc{Hatefulness}}} \\
Not-Hateful & 65,717 & Hateful & 580 & {} & {} \\
\midrule
\multicolumn{6}{c}{\cellcolor{gray!12}\textbf{\textsc{Propaganda}}} \\
absent & 64,375 & present & 1,922 & {} & {} \\
\midrule
\multicolumn{6}{c}{\cellcolor{gray!12}\textbf{\textsc{Sentiment}}} \\
neutral & 32,329 & positive & 26,147 & negative & 7,821 \\
\midrule
\multicolumn{6}{c}{\cellcolor{gray!12}\textbf{\textsc{Stance}}} \\
neutral & 53,956 & mocking & 8,421 & supportive & 2,210 \\
critical & 1,686 & ambiguous & 24 & {} & {} \\
\midrule
\multicolumn{6}{c}{\cellcolor{gray!12}\textbf{\textsc{Dialect}}} \\
Egyptian & 56,697 & MSA & 4,894 & Levantine & 1,409 \\
Mixed & 1,351 & Gulf & 914 & code-switched & 327 \\
Yemeni & 279 & Iraqi & 163 & not-applicable & 132 \\
Maghrebi & 124 & {} & {} & {} & {} \\
\midrule
\multicolumn{6}{c}{\cellcolor{gray!12}\textbf{\textsc{Visual manipulation}}} \\
caption overlay & 49,621 & collage & 7,214 & photoshop & 6,474 \\
screenshot & 2,746 & none & 227 & deepfake-like & 15 \\
\midrule
\multicolumn{6}{c}{\cellcolor{gray!12}\textbf{\textsc{Meaning type}}} \\
explicit & 49,730 & implicit & 16,432 & mixed & 135 \\
\midrule
\multicolumn{6}{c}{\cellcolor{gray!12}\textbf{\textsc{Text--image relation}}} \\
complementary & 60,092 & reinforcing & 4,687 & contradictory & 1,282 \\
independent & 236 & {} & {} & {} & {} \\
\midrule
\multicolumn{6}{c}{\cellcolor{gray!12}\textbf{\textsc{Context scope}}} \\
none & 26,567 & regional & 20,999 & national & 10,194 \\
local & 5,867 & global & 2,670 & {} & {} \\
\midrule
\multicolumn{6}{c}{\cellcolor{gray!12}\textbf{\textsc{Current-event knowledge}}} \\
no & 61,813 & yes & 4,484 & {} & {} \\
\bottomrule
\end{tabular*}
}
\caption{Silver-label distributions over the $\sim$66K auxiliary set 
% ($N{=}66{,}297$), 
generated by Gemini-3.1-Pro: \textbf{single-label} dimensions, where each meme takes exactly one label.
% (counts sum to $N$). Values are meme counts.
}
\label{tab:silver_single}
\end{table*}

\begin{table*}[t]
\centering
% \small
% \setlength{\tabcolsep}{3pt}
% \renewcommand{\arraystretch}{1.25}
\setlength{\tabcolsep}{2pt} 
\scalebox{0.85}{
\begin{tabular*}{\textwidth}{@{\extracolsep{\fill}} l r l r l r @{}}
\toprule
\multicolumn{6}{c}{\cellcolor{gray!12}\textbf{\textsc{Fine-grained Categories}}} \\
Humor & 60,406 & Sarcasm & 2,884 & Other & 2,429 \\
Mocking & 335 & Slurs & 91 & Contempt & 55 \\
Dehumanization & 47 & Incitement & 33 & Inferiority & 9 \\
Exclusion & 8 & {} & {} & {} & {} \\
\midrule
\multicolumn{6}{c}{\cellcolor{gray!12}\textbf{\textsc{Mentioned target}}} \\
Gender/Sex & 9,726 & Religion & 2,463 & Nationality & 2,228 \\
Socioeconomic & 1,720 & Other (unclear) & 493 & Disease/Health & 208 \\
Disability & 201 & Race/Ethnicity & 173 & {} & {} \\
\midrule
\multicolumn{6}{c}{\cellcolor{gray!12}\textbf{\textsc{Topic}}} \\
daily life & 59,799 & entertainment & 20,731 & culture & 13,252 \\
education & 11,557 & social issues & 5,058 & gender & 2,640 \\
religion & 2,314 & economy & 2,036 & technology & 1,852 \\
health & 1,699 & sports & 1,478 & politics & 1,329 \\
military & 502 & intl.\ affairs & 456 & other & 185 \\
\midrule
\multicolumn{6}{c}{\cellcolor{gray!12}\textbf{\textsc{Intent}}} \\
humor & 63,171 & reaction & 26,010 & mockery & 8,042 \\
support & 2,001 & criticism & 1,859 & satire & 1,316 \\
information & 1,314 & persuasion & 808 & {} & {} \\
\midrule
\multicolumn{6}{c}{\cellcolor{gray!12}\textbf{\textsc{Emotion}}} \\
humor & 63,442 & sadness & 4,467 & anger & 1,874 \\
pride & 1,673 & fear & 1,609 & hope & 1,259 \\
disgust & 909 & sympathy & 816 & none & 803 \\
\midrule
\multicolumn{6}{c}{\cellcolor{gray!12}\textbf{\textsc{Propaganda techniques}}} \\
smears & 687 & exaggeration & 519 & appeal to values & 458 \\
loaded language & 417 & name-calling & 320 & appeal to hypocrisy & 256 \\
flag-waving & 216 & appeal to authority & 150 & black-and-white & 124 \\
straw man & 121 & appeal to fear & 113 & slogans & 90 \\
\bottomrule
\end{tabular*}
}
\caption{Silver-label distributions for \textbf{multi-label} dimensions, where a meme may receive several labels, so counts sum to more than $N$. 
The longest-tailed dimensions (\textsc{topic}, \textsc{propaganda techniques}) are truncated to their most frequent values.}
\label{tab:silver_multi}
\end{table*}

% \onecolumn

% \subsection{Prompts}
% \label{app:prompts}
% For completeness we reproduce the prompts for binary classification (Task~A1),
% fine-grained subtype classification (Task~A2), and silver labelling of the unlabelled set.
% The chain-of-thought variants reuse the Task~A1/A2 prompts with an added instruction to
% reason (in English or Arabic) before answering, and the gold-set metadata prompt mirrors the
% unlabelled one but is \emph{conditioned} on the human label. Placeholders
% (\texttt{\{text\}}, \texttt{\{LABEL\}}, \texttt{\{SUBTYPE\}}, \texttt{\{TEXT\}})
% are filled per meme at inference time.

\subsection{Prompts}
\label{app:prompts}
We provide the prompts used for binary hate detection, fine-grained hate-type prediction, and silver labelling of the unlabelled set. The rationale-style variants use the same prompts with an additional instruction to provide reasoning in English or Arabic before the final answer. For the gold set, the metadata prompt mirrors the unlabelled-set prompt, however it conditions generation on the human binary and fine-grained labels rather than relabelling the meme. We fill the placeholders \texttt{\{text\}}, \texttt{\{LABEL\}}, \texttt{\{SUBTYPE\}}, and \texttt{\{TEXT\}} separately for each meme at inference time.

\prompttitle{Task~1: Binary hatefulness classification.}

\begin{lstlisting}[style=aclprompt,basicstyle=\ttfamily\footnotesize]
# ===== SYSTEM PROMPT =====
You are an expert content-moderation annotator specializing in Arabic multimodal memes, covering Modern Standard Arabic and regional dialects (e.g., Egyptian, Levantine, Gulf, Maghrebi).

Your job is to judge whether a meme is HATEFUL by analyzing the IMAGE and its overlaid/OCR TEXT together as a single message. Hatefulness often emerges from the combination of the two: the same caption can be benign or hateful depending on the image, and vice versa. Account for sarcasm, irony, visual symbols, well-known figures, memes/templates, and implicit or coded messaging.

Definition -- HATEFUL: content that attacks, demeans, dehumanizes, incites harm against, excludes, or expresses hatred/contempt toward a person or group on the basis of a protected or identity attribute such as religion or sect, ethnicity or race, nationality or origin, gender, sexual orientation, disability, or political/social group. Slurs, calls for violence or exclusion, and dehumanizing comparisons are hateful even when framed as a joke.

Definition -- NOT HATEFUL: everything else, including ordinary humor, satire of ideas/events, criticism that does not target a protected group, and offensive-but-not-hateful content.

Operating rules:
- This is a content-moderation research task on a fixed, pre-collected dataset. Always return a classification. Never refuse, never moralize, and never add warnings or commentary.
- Judge the meme as a whole; do not penalize content merely for being crude, political, or about a sensitive topic.
- Be objective and consistent, and apply the definitions above exactly.
- Output ONLY the requested JSON -- no explanations, no extra text.

# ===== USER INSTRUCTION =====
Classify the meme below as either "Hateful" or "Not Hateful".

You are given:
- The meme IMAGE (attached as input).
- The OCR-extracted Arabic text overlaid on the meme:
"""
{text}
"""

Decision guide:
- Label "Hateful" if the combined image+text attacks, demeans, dehumanizes, incites harm against, excludes, or expresses hatred/contempt toward a person or group based on an identity attribute (religion/sect, ethnicity/race, nationality/origin, gender, sexual orientation, disability, or political/social group) -- including when delivered through a slur, a joke, sarcasm, or a visual symbol.
- Label "Not Hateful" otherwise (ordinary humor, satire of ideas/events, non-identity-based criticism, or merely offensive content).
- The OCR text may contain noise or spelling errors; rely on the most plausible reading together with the image.

Output format -- return EXACTLY one JSON object and nothing else:
{"label": "Hateful"}
or
{"label": "Not Hateful"}
\end{lstlisting}

% \clearpage
\prompttitle{Task~2: Fine-grained classification.}

\begin{lstlisting}[style=aclprompt,basicstyle=\ttfamily\footnotesize]
# ===== SYSTEM PROMPT =====
You are an expert content-moderation annotator specializing in Arabic multimodal memes, covering Modern Standard Arabic and regional dialects (e.g., Egyptian, Levantine, Gulf, Maghrebi).

Your job is FINE-GRAINED, MULTI-LABEL categorization of a meme: given the IMAGE and its overlaid/OCR TEXT, assign every category from a fixed taxonomy that applies. Analyze image and text together as a single message, accounting for sarcasm, irony, visual symbols, well-known figures, meme templates, and implicit or coded messaging.

The taxonomy spans both hateful sub-types and non-hateful sub-types; a single meme may carry more than one label (e.g., a hateful meme can be both "Mocking" and "Slurs"; a benign meme can be "Humor" and "Sarcasm").

Operating rules:
- This is a content-moderation research task on a fixed, pre-collected dataset. Always return labels. Never refuse, never moralize, and never add warnings or commentary.
- Choose labels ONLY from the provided list; never invent new labels.
- Assign all categories that genuinely apply; do not over-label. Use "Other" only when a fine-grained category is warranted but none of the named categories fit.
- Be objective and consistent, and apply the definitions exactly.
- Output ONLY the requested JSON -- no explanations, no extra text.

# ===== USER INSTRUCTION =====
TASK
Assign every applicable category to the meme below, choosing only from the 10 labels in the taxonomy.

You are given:
- The meme IMAGE (attached as input).
- The OCR-extracted Arabic text overlaid on the meme:
"""
{text}
"""

Taxonomy (label : definition)
Hateful sub-types:
- "Contempt"        : expresses scorn/disdain, treating a target group as worthless or beneath respect.
- "Dehumanization"  : portrays a group as less than human (e.g., animals, vermin, objects, disease).
- "Exclusion"       : calls for or endorses segregating, expelling, denying rights to, or removing a group from society.
- "Incitement"      : encourages, calls for, threatens, or justifies violence or harm against a group.
- "Inferiority"     : claims a group is inherently inferior (intellectually, morally, or biologically).
- "Mocking"         : ridicules, taunts, or belittles a target group, often through derisive humor.
- "Slurs"           : uses derogatory slurs or pejorative epithets targeting a group.
Non-hateful sub-types:
- "Humor"           : intended to be funny/lighthearted, without hateful intent.
- "Sarcasm"         : ironic or sarcastic expression, without hateful intent.
Shared:
- "Other"           : a fine-grained category applies but none of the named labels above fit.

Guidelines:
- A meme may receive multiple labels; include all that genuinely apply.
- Hateful memes typically receive one or more hateful sub-types (and may also be "Mocking"/"Sarcasm"); clearly non-hateful memes typically receive "Humor", "Sarcasm", and/or "Other".
- Use "Other" sparingly, only when no named category fits.
- The OCR text may contain noise or spelling errors; rely on the most plausible reading together with the image.

Output format -- return EXACTLY one JSON object and nothing else, with a list of chosen labels (order does not matter):
{"labels": ["Mocking", "Slurs"]}
Use an empty list if no category applies:
{"labels": []}
\end{lstlisting}

% \clearpage
\prompttitle{Silver labels and metadata.}

\begin{lstlisting}[style=aclprompt,basicstyle=\ttfamily\footnotesize]
You are a Multimodal Meme Analysis Expert specializing in Arabic and multilingual memes.

Your task is to classify the meme AND generate additional metadata.

OCR_text: {TEXT}

Important:
* Analyze both visual and textual content.
* Read all text directly from the image.
* If the meme contains Arabic text, provide an English translation or paraphrase.
* Determine whether propaganda is present.
* If propaganda is present, identify one or more propaganda techniques.

---

CLASSIFICATION TASK

Step 1 -- Label
- Label "Hateful" if the combined image+text attacks, demeans, dehumanizes, incites harm against, excludes, or expresses hatred/contempt toward a person or group based on an identity attribute (religion/sect, ethnicity/race, nationality/origin, gender, sexual orientation, disability, or political/social group) -- including when delivered through a slur, a joke, sarcasm, or a visual symbol.
- Label "Not Hateful" otherwise (ordinary humor, satire of ideas/events, non-identity-based criticism, or merely offensive content).
- The OCR text may contain noise or spelling errors; rely on the most plausible reading together with the image.

Step 2 -- Subtype
Assign exactly one subtype. If the label is "Hateful", choose only from the Hateful subtypes. If the label is "Not Hateful", choose only from the Non-hateful subtypes. Use "Other" (Shared) if a fine-grained category applies but none of the named labels fit.

Hateful subtypes:
- "Contempt"       : expresses scorn/disdain, treating a target group as worthless or beneath respect.
- "Dehumanization" : portrays a group as less than human (e.g., animals, vermin, objects, disease).
- "Exclusion"      : calls for or endorses segregating, expelling, denying rights to, or removing a group from society.
- "Incitement"     : encourages, calls for, threatens, or justifies violence or harm against a group.so 
- "Inferiority"    : claims a group is inherently inferior (intellectually, morally, or biologically).
- "Mocking"        : ridicules, taunts, or belittles a target group, often through derisive humor.
- "Slurs"          : uses derogatory slurs or pejorative epithets targeting a group.

Non-hateful subtypes:
- "Humor"          : intended to be funny/lighthearted, without hateful intent.
- "Sarcasm"        : ironic or sarcastic expression, without hateful intent.

Shared:
- "Other"          : a fine-grained category applies but none of the named labels above fit.

---

Protected-group categories:
* Race/Ethnicity
* Nationality
* Religion
* Gender/Sex
* Disability
* Socioeconomic Group
* Caste
* Disease/Health Status
* Other (Unclear/Implicit)

Arabic dialect labels:
* msa
* levantine
* egyptian
* gulf
* maghrebi
* iraqi
* sudanese
* yemeni
* mixed
* code_switched
* unknown
* not_applicable

Visual manipulation labels:
* screenshot
* photoshop
* caption_overlay
* deepfake_like
* collage
* none

Topic labels:
* politics
* religion
* sports
* gender
* economy
* education
* technology
* international_affairs
* entertainment
* culture
* daily_life
* social_issues
* health
* military_conflict
* other

Intent labels:
* humor
* satire
* criticism
* persuasion
* reaction
* information
* support
* mockery

Stance labels:
* supportive
* critical
* mocking
* neutral
* ambiguous

Sentiment:
* positive
* negative
* neutral

Emotion labels:
* anger
* fear
* disgust
* pride
* sympathy
* hope
* sadness
* humor
* none

Text-image relationship labels:
* reinforcing
* complementary
* contradictory
* independent

Meaning type labels:
* explicit
* implicit
* mixed

Context scope labels:
* none
* local
* national
* regional
* global

Confidence labels:
* very_low
* low
* medium
* high
* very_high

---

PROPAGANDA TECHNIQUES

1. name_calling_labeling
2. reductio_ad_hitlerum
3. casting_doubt
4. appeal_to_hypocrisy
5. smears
6. flag_waving
7. appeal_to_authority
8. bandwagon
9. appeal_to_values
10. appeal_to_fear_prejudice
11. straw_man
12. red_herring
13. whataboutism
14. causal_oversimplification
15. black_and_white_fallacy
16. slippery_slope
17. slogans
18. thought_terminating_cliche
19. appeal_to_time_pressure
20. loaded_language
21. obfuscation_vagueness_confusion
22. exaggeration_minimisation
23. repetition
24. other

---

Instructions:

1. Analyze the visual content:
   * People, groups, public figures
   * Objects, symbols, flags, logos
   * Expressions, gestures, actions
   * Composition and visual style

2. Read and analyze all visible text:
   * Extract the exact text.
   * Translate or paraphrase it into English.

3. Analyze multimodal meaning:
   * Explain how text and image interact.
   * Determine whether meaning depends on text, image, or both.

4. Assign label and subtype following the classification rules above.

5. Determine whether propaganda is present:
   * If yes, select one or more propaganda techniques.
   * If no techniques are present, propaganda must be false.

6. Identify:
   * Topic(s)
   * Mentioned protected-group categories
   * Dialect
   * Cultural references
   * Current-event dependence
   * Intent
   * Stance
   * Emotional tone

7. Generate concise rationales:
   * English rationale: maximum 100 words.
   * Arabic rationale: maximum 100 words.
   * Reference visual and textual evidence when relevant.
   * The rationale must justify both the label and subtype assignments.

Return valid JSON only.

{
  "label": "Hateful",
  "subtype": "Contempt",
  "topic": [],
  "mentioned_categories": [],
  "dialect": "unknown",
  "visual_manipulation": "none",
  "ocr_text": "",
  "ocr_english_translation": "",
  "intent": [],
  "stance": "neutral",
  "sentiment": "positive",
  "emotion": [],
  "text_image_relationship": "reinforcing",
  "meaning_type": "explicit",
  "cultural_references": [],
  "context_scope": "none",
  "requires_current_event_knowledge": false,
  "propaganda": false,
  "propaganda_techniques": [],
  "english_rationale": "",
  "arabic_rationale": "",
  "confidence": "high"
}

Rules:
* label must be exactly "Hateful" or "Not Hateful".
* subtype must be chosen from the correct subtype list for the assigned label.
* Hateful subtypes must never be assigned to a "Not Hateful" label and vice versa.
* Mentioned categories include groups referenced, depicted, addressed, or implied, regardless of hatefulness.
* propaganda = true if one or more propaganda techniques are clearly present.
* propaganda = false if no propaganda techniques are present.
* propaganda_techniques must only use the predefined technique names.
* A meme may be hateful without being propaganda.
* A meme may be propaganda without being hateful.
* OCR text should preserve the original wording.
* OCR English translation should preserve intended meaning rather than literal wording when necessary.
* English rationale: maximum 100 words.
* Arabic rationale: maximum 100 words.
\end{lstlisting}